\documentclass[11pt]{article}

\usepackage[preprint]{acl}

\usepackage{times}
\usepackage{latexsym}

\usepackage[T1]{fontenc}

\usepackage[utf8]{inputenc}

\usepackage{microtype}

\usepackage{inconsolata}

\usepackage{graphicx}

\usepackage{hyperref}
\usepackage{url}

\usepackage[utf8]{inputenc} 
\usepackage[T1]{fontenc}    
\usepackage{hyperref}       
\usepackage{url}            
\usepackage{booktabs}       
\usepackage{amsfonts}       
\usepackage{nicefrac}       
\usepackage{microtype}      
\usepackage{xcolor}         
\usepackage{natbib}

\usepackage{soul}
\usepackage{amsmath}
\usepackage{graphicx}
\usepackage{array}     

\usepackage{multirow}

\usepackage{adjustbox}
\usepackage{xcolor}
\usepackage[table]{xcolor}
\usepackage{subcaption}
\usepackage{pifont}
\usepackage{enumitem}

\usepackage{xspace}
\usepackage[normalem]{ulem}

\usepackage{wrapfig}
\usepackage{lipsum}

\usepackage{multirow}
\usepackage{listings}
\usepackage{pifont}

\usepackage{amssymb} 

\usepackage{makecell}

\usepackage{dblfloatfix}

\definecolor{gainray}{gray}{0.30} 
\definecolor{dropray}{gray}{0.70} 

\newcommand{\gn}[1]{\textcolor{gainray}{#1}} 
\newcommand{\dpn}[1]{\textcolor{dropray}{#1}} 

\newcommand{\eg}{\emph{e.g.}\xspace}

\title{Breaking Down and Building Up: Mixture of Skill-Based Vision-and-Language Navigation Agents}

\author{
  Tianyi Ma\textsuperscript{1} \quad
  Yue Zhang\textsuperscript{1} \quad
  Zehao Wang\textsuperscript{2} \quad
  Parisa Kordjamshidi\textsuperscript{1} \\
  \\
  \textsuperscript{1}Michigan State University
  \textsuperscript{2}ESAT-PSI, KU Leuven \\
  \href{mailto:matiany3@msu.edu}{matiany3@msu.edu},
  \href{mailto:kordjams@msu.edu}{kordjams@msu.edu}
}

\begin{document}
\maketitle

\renewcommand{\thefootnote}{}
\footnotetext{Preprint.}
\renewcommand{\thefootnote}{\arabic{footnote}}
\setcounter{footnote}{0}

\begin{abstract}

Vision-and-Language Navigation (VLN) poses significant challenges for agents to interpret natural language instructions and navigate complex 3D environments. 
While recent progress has been driven by large-scale pre-training and data augmentation, current methods still struggle to generalize to unseen scenarios, particularly when complex spatial and temporal reasoning is required. 
In this work, we propose SkillNav\footnote{Project page: \href{https://hlr.github.io/SkillNav/}{https://hlr.github.io/SkillNav/} \quad Code: \href{https://github.com/HLR/SkillNav}{https://github.com/HLR/SkillNav}}, a modular framework that introduces structured, skill-based reasoning into Transformer-based VLN agents. Our method decomposes navigation into a set of interpretable atomic skills (\textit{e.g.}, Vertical Movement, Area and Region Identification, Stop and Pause), each handled by a specialized agent. 
To support targeted skill training without manual data annotation, we construct a synthetic dataset pipeline that generates diverse, linguistically natural, skill-specific instruction-trajectory pairs.
We then introduce a novel training-free Vision-Language Model (VLM)-based router, which dynamically selects the most suitable agent at each time step by aligning sub-goals with visual observations and previous actions. 
SkillNav obtains competitive results on commonly used benchmarks and establishes state-of-the-art generalization on GSA-R2R, a benchmark with novel instruction styles and unseen environments. 

\end{abstract}
\section{Introduction}\label{sec:introduction}
\begin{figure}[ht]
    \centering    \includegraphics[width=\linewidth]{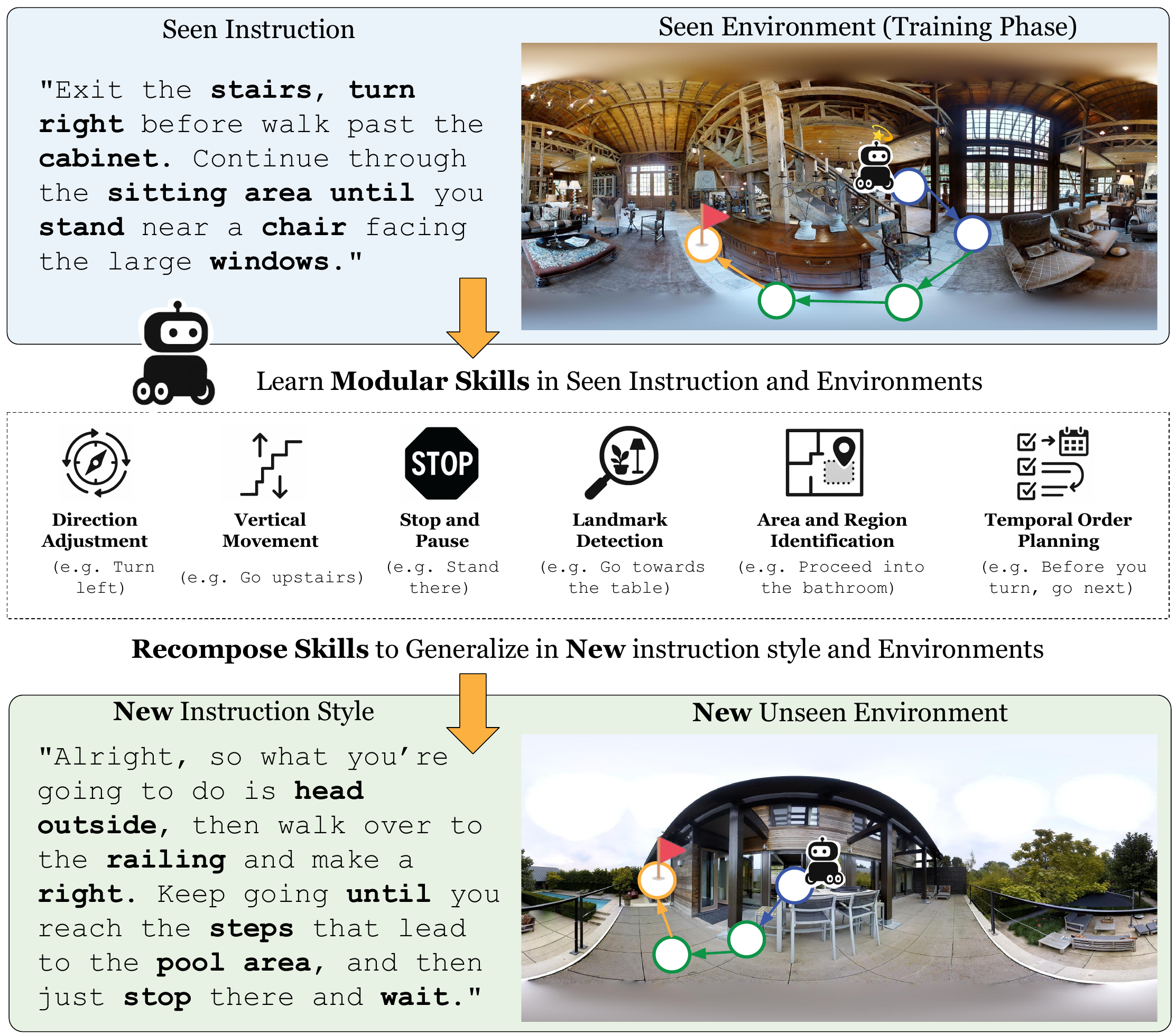}
    \caption{SkillNav decomposes complex navigation instructions into atomic skills, which can be flexibly recomposed to address new environments.}
    \label{fig:motivation_decomposition}
\end{figure}
Vision-and-Language Navigation (VLN)~\citep{anderson2018vision, zhang2024visionandlanguage} is a  subfield of embodied AI that integrates natural language understanding, visual perception, and sequential decision-making to allow autonomous agents to navigate and interact within visual environments. 
With the rise of foundation models~\citep{zhou2024comprehensive, xiao2025foundations, li2024multimodal, zhang2024vision}, VLN has seen notable progress in multimodal grounding and generalization.

Despite recent advances, a key challenge in VLN lies in enabling agents to generalize reliably and interact with unseen environments and novel instructions. 
Previous approaches have enhanced VLN agents' generalization ability through extensive
training on large-scale synthetic instruction-trajectory pairs across varied environments~\citep{hao2020prevalent, Chen_2022_HM3D_AutoVLN, wang2023scaling, wang_bootstrapping_2024}.
While data-driven methods improve VLN agents' generalization, their main limitation is reliance on black-box, end-to-end models~\citep{anderson2018vision, hong_recurrent_2021} that tend to memorize training examples. This restricts their effectiveness in unobserved scenarios requiring deeper compositional reasoning, such as understanding diverse instructions, temporal relationships, or complex landmarks, and generalizing across a wide range of visual environments.
%
%
Beyond data-intensive supervised approaches, recent work has explored zero-shot approaches leveraging Large Language Models (LLMs) for VLN tasks to improve generalization ability~\citep{zhou_navgpt_2023, long2024discuss, chen_mapgpt_2024, zhang2025flexvln}. Although zero-shot LLM-based agents show relatively stable performance across seen and unseen environments, they still considerably lag behind fine-tuned VLN models. 
Specifically, we observe a significant performance gap ($\sim36\%$ in Success Rate),
primarily arising from intrinsic limitations of LLMs, including their insufficient grounding in embodied environments and imprecise alignment of linguistic instructions with specific navigational actions.
This gap highlights the urgent need for methods capable of bridging the broad generalization and compositional reasoning of LLMs with precise, visually grounded execution.

To address these limitations, we propose \textbf{SkillNav}, a modular VLN framework that decomposes navigation learning into individual and reusable skills, enabling flexible re-composition and enhanced generalization in new environments as shown in Figure~\ref{fig:motivation_decomposition}. Unlike prior methods that treat instruction execution as an end-to-end mapping from instructions directly to actions, SkillNav explicitly captures the compositional nature of navigation tasks. Furthermore, we introduce a novel Vision-Language Model (VLM)-based router that leverages multi-modal reasoning to dynamically select the most appropriate skill at each navigation step, conditioned on the current sub-instruction, visual observation, and historical actions. 
SkillNav not only improves interpretability by making the decision-making processes more transparent but also facilitates robust adaptation to diverse instructions and unseen visual environments.

Specifically, we build on previous research~\citep{wang-etal-2024-navigating}, and identify a set of atomic skills required for effectively completing the VLN task. 
For each skill, we construct a dataset containing relevant instructions paired with corresponding visual observations, and fine-tune a dedicated agent on top of a strong VLN backbone. This process yields five specialized skill agents, covering Direction Adjustment, Vertical Movement, Stop and Pause, Landmark Detection, and Area and Region Identification,  each proficient in its designated capability. 
%
To orchestrate these specialized agents for complex navigation tasks, we introduce a VLM-based Action Router that operates in three distinct phases. First, a Temporal Reordering Module processes the natural language instruction to generate a chronologically ordered sequence of subgoals, resolving implicit temporal dependencies. Next, a Subgoal Localizer dynamically identifies the immediate objective by aligning the reordered plan with the agent's visual and action history. Finally, a Skill Router analyzes the active subgoal and context to select the most suitable expert agent to execute the next navigation action.



SkillNav attains a strong performance on the Room-to-Room (R2R) benchmark~\citep{anderson2018vision}, and 
achieves state-of-the-art (SOTA) generalization on the GSA-R2R benchmark~\citep{hong_general_2025}. GSA-R2R contains novel dialogue-style instructions and diverse visual environments with both unseen residential and non-residential settings. 
%
Additionally, we evaluate individual skill-based agents using NavNuances~\citep{wang-etal-2024-navigating}, a dataset specifically designed for fine-grained skill evaluation.
%
%
We report ablations and qualitative analysis to isolate each component's contribution and motivate the router design.
Our contributions are summarized as follows:
\begin{enumerate}[leftmargin=1.2em, itemsep=2pt, topsep=2pt, parsep=0pt, partopsep=0pt]


    \item We propose \textbf{SkillNav}, a modular framework that decomposes navigation tasks into atomic skills to combine VLM reasoning with specialized fine-tuned agents, significantly enhancing generalization capabilities to novel instructions and unseen visual environments.
    
    
    \item We design a synthetic data generation pipeline that creates skill-specific but linguistically diverse supervision without requiring expensive and labor-intensive human annotation.
    
    \item We demonstrate state-of-the-art generalization on the challenging GSA-R2R dataset and validate our framework through a comprehensive ablation study.

\end{enumerate}
\section{Related Work}

\noindent\textbf{Vision-and-Language Navigation Models.}
A wide range of methods have been proposed for addressing VLN tasks. These methods have evolved from early LSTM-based architectures~\citep{anderson2018vision, tan2019learning}
to Transformer-based models~\citep{chen2021history, chen_think_2022, an2023bevbert}
and, most recently, to Large Language Model (LLM)-based agents~\citep{zhou_navgpt_2023, chen_mapgpt_2024, lin_navcot_2024, zhou_navgpt-2_2024, zheng2024towards_navillm, zhang-2025-vln-analogical, zhang2025spartund}.
A primary challenge remains improving generalization to unfamiliar environments.
%
Existing methods largely rely on data augmentation strategies for visual observations~\citep{li2022envedit, liu2021vision, li2023panogen} or instructions~\citep{wang2023scaling, wang_bootstrapping_2024, hao2020prevalent, zhang-kordjamshidi-2023-vln-trans, zhang-etal-2024-navhint}.
However, these end-to-end supervised approaches often memorize training trajectories rather than mastering the compositional reasoning required in novel or unseen scenarios.
While recent zero-shot LLM agents~\citep{zhou_navgpt_2023, chen_mapgpt_2024, long2024discuss, zhang2025flexvln} avoid task-specific training, they lack of detailed spatial understanding and precise grounding in action execution.
%
In contrast, we propose SkillNav, a modular framework that explicitly decomposes VLN tasks into reusable navigation skills. Each skill is individually fine-tuned for precise spatial grounding, while high-level reasoning and flexible skill composition leverage LLMs and VLMs, significantly improving generalization to unseen environments and varied instructions.
\noindent\textbf{Multi-Agent Collaboration in Vision-Language Navigation.}
Recent multi-agent VLN work has moved toward debate-style systems like DiscussNav~\citep{long2024discuss} and planner executor hierarchies like FlexVLN~\citep{zhang2025flexvln} and CLASH~\citep{wang2025clash}. 
Although these approaches can generalize better, they typically rely on ensembles that activate multiple models per step, incurring structural redundancy. When these generalist ensembles conflict, systems often default to ungrounded zero-shot LLM decisions, sacrificing in-domain precision for broader adaptability.
%
SkillNav overcomes this dilemma by decomposing navigation into a library of specialized skill agents. A VLM router selects one best-fit specialist for the current sub-goal instead of invoking an ensemble.
This enables the VLM to perform compositional planning by decomposing complex instructions into manageable sub-goals and mapping them to specific skills, while specialized agents ensure precise, grounded action execution.
Consequently, SkillNav achieves state-of-the-art generalization on unfamiliar benchmarks like GSA-R2R while remaining competitive on in-domain datasets such as R2R, effectively balancing computational efficiency with robust performance.
\section{Preliminaries}\label{method: preliminary}

In the VLN problem setting, an agent navigates through an environment by following a natural language instruction $I$ to reach a specified target location. The environment is discretized into a connectivity graph $\mathcal{G} = (V, E)$,
%
where $V$ denotes a non-empty set of navigable nodes,
and $E$ is a set of undirected connectivity edges. At each time step $t$, the agent located at viewpoint $v_t$ receives a panorama represented by $n$ images, denoted as $D_{t}=\{o_{i}\}_{i=0}^{n}$. The agent is aware of a subset of views $O_t \subseteq D_{t}$ heading towards its navigable neighboring nodes $\mathcal{N}(v_{t})$. The local action space $A_t$ contains navigating to node $v \in \mathcal{N}(v_{t})$ or stopping at current node $v_{t}$.

In this work, we leverage DUET~\citep{chen_think_2022} as our base VLN agent. DUET is a graph transformer-based architecture designed to balance long-term planning on a topological map with fine-grained visual grounding of the current panoramic view. 
To integrate this architecture into our framework, we formulate its policy as:
\begin{equation}
\label{eq:duet}
    a^{*}_{t} = \pi(I, O_t, M_t).
\end{equation}
where $M_t \subseteq \mathcal{G}$ denotes the online constructed topological map observed after $t$ steps of navigation, and $a^{*}_{t} \in A_{t}$ is the predicted action.

\begin{figure*}[t]
    \centering
    \includegraphics[width=\linewidth]{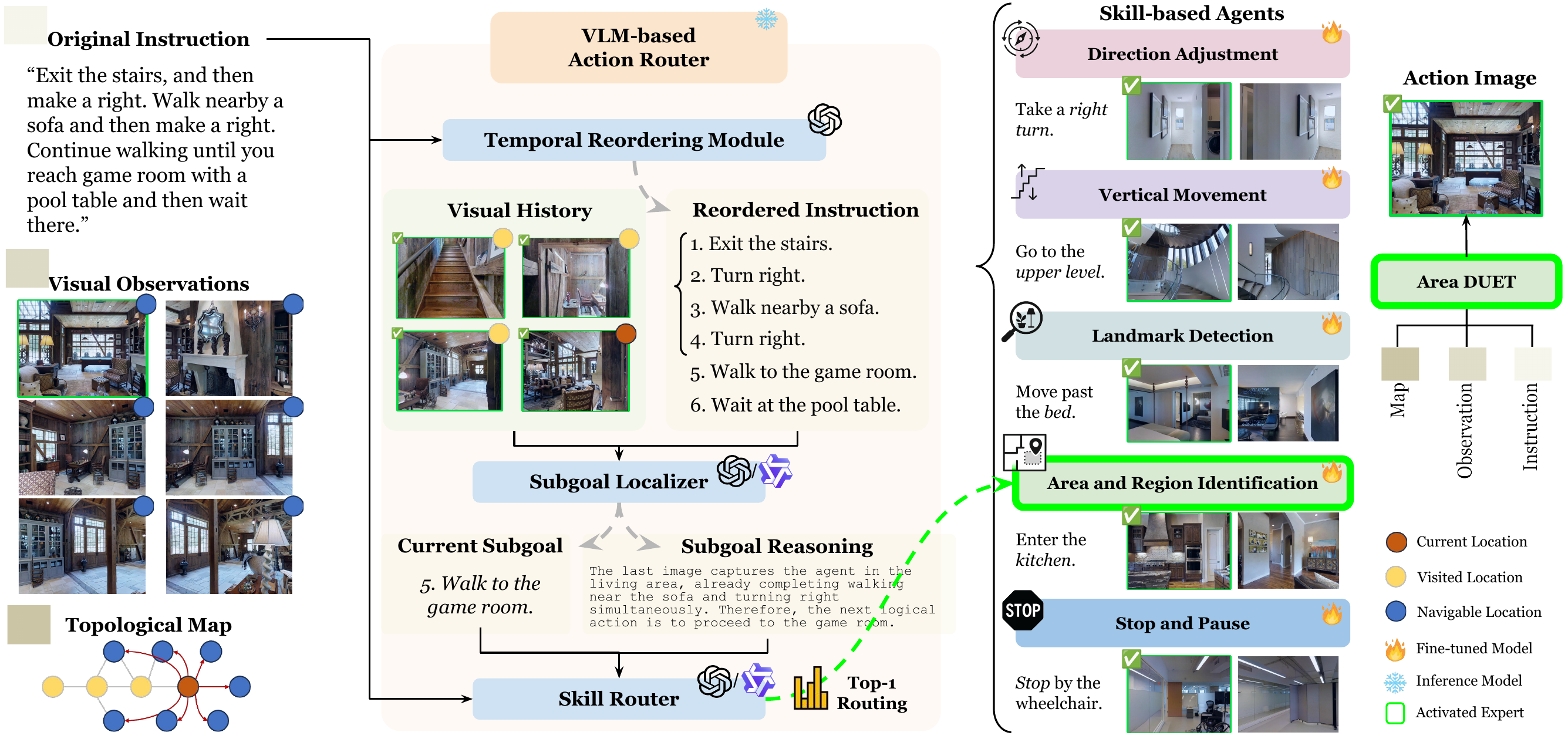}
    \caption{\textbf{SkillNav architecture.} Inputs are visual observations, the original instruction, and the online topological map. The VLM-based action router (i) applies temporal reordering to produce an ordered list of subgoals (used solely to assist subgoal localization), (ii) localizes the next subgoal using visual and subgoal history based on the reordered instruction, and (iii) routes to the most suitable skill-based agent. The chosen agent conditions on the original instruction, current observations, and map to output the navigation action.
    }

    \label{fig:architecture}
\end{figure*}

\section{Methodology}


We propose \textbf{SkillNav}, a framework that coordinates a set of agents, each trained for a specific atomic skill, to solve the navigation problem.
SkillNav enhances generalization by treating navigation as a composition of atomic skills rather than a direct language-to-action mapping. This design mirrors how humans transfer sub-skills across unfamiliar situations, preventing overfitting to complex trajectories and enabling systematic reuse of skills across environments and instruction styles.
As shown in Figure~\ref{fig:architecture}, the framework comprises two key components: a VLM-based router that performs temporal reordering, subgoal localization, and skill selection, and a set of skill-specific agents.
%
Each agent is built upon the DUET architecture and trained with tailored synthetic data to make skill-conditioned decisions, as detailed in this section.
%

\subsection{Skill Taxonomy}\label{sec:skill_taxonomy}

We use the skills defined in NavNuances~\citep{wang-etal-2024-navigating} that appear to be essential for building a robust VLN agent. NavNuances provides skill categories and creates a diagnostic dataset to analyze models' errors. However, it does not provide solutions for improving the agent's skills. In this work, we extend the initially proposed skill categories and provide solutions for acquiring them by skill-based agents.

%
%
We adopt four frequently observed atomic skills from NavNuances, \textbf{Direction Adjustment}, \textbf{Vertical Movement}, \textbf{Landmark Detection}, and \textbf{Area and Region Identification}. Detailed NavNuances skill definitions and annotation cues are provided in Appendix~\ref{sec:navnuances_atomic_skill_def}.
Moreover, we find persistent challenges in temporal reasoning and stop criteria. Errors in temporal reasoning often disrupt the correct order of subgoal execution. Critical stop decisions are sometimes made too early or too late, reducing navigation success. 
To address these issues, we extend the skill taxonomy with two additional skills: \textbf{Stop and Pause} and \textbf{Temporal Order Planning}.
In the following, we elaborate on these two new skills and their roles in navigation.

\textbf{Stop and Pause} captures the agent's ability to dynamically control motion termination and temporary halting in response to visual or linguistic cues. This includes recognizing explicit stop commands (\eg, ``Stop at the doorway'') or context-sensitive halts triggered by landmarks or obstacles (\eg, ``Pause when you see the red sign''). 
The stop and pause skill emphasizes precise temporal-spatial control to ensure safe, context-aware navigation.

\textbf{Temporal Order Planning} reflects the agent's capability to reason over the sequence and structure of subgoals. This includes understanding conditional immediacy (\eg, ``Once you enter the hallway, turn left''), maintaining actions for a bounded duration (\eg, ``Keep walking until you see the staircase''), executing forward sequential steps (\eg, ``Go forward, then turn right, and finally stop''), and handling backward references to prior states (\eg, ``Before turning, make sure you're at the hallway entrance''). Effective temporal order planning involves temporal relations that guide both when and how atomic skills should be executed.



To quantify the presence and frequency of these skills in R2R~\citep{anderson2018vision}, we perform a keyword-based analysis of the navigation instructions as shown in Figure~\ref{fig:skill_instruction_distribution} in Appendix~\ref{sec:data_synthesis_appendix}.
Each instruction is scanned for a curated set of indicative keywords, compiled for each skill category based on linguistic patterns observed in prior datasets and real-world navigation discourse. 
For instance, terms like ``wait'' or ``stay'' are used to detect Stop and Pause,
%
while words such as ``stairs'' or ``elevator'' signal Vertical Movement. An instruction can be counted for multiple skills if it exhibits multiple relevant keywords. 
%

This taxonomy is intended as a task-grounded and extensible skill library rather than a universally complete inventory. The selected skills correspond to frequent linguistic and geometric requirements in indoor VLN, and Appendix~\ref{sec:appendix_skill_taxonomy_sensitivity} further reports skill definitions, R2R coverage, and sensitivity analysis in Table~\ref{tab:skill_set_sensitivity} showing that the five-skill decomposition is optimal for delegation precision. New environments can extend the same framework by adding a specialized skill executor and exposing its definition to the router, without changing the high-level routing mechanism.

\subsection{Skill-Specific Data Synthesis and Agent Training}\label{sec:data_synthesis_agent_train}

\begin{table}[t]
    \centering
    \caption{Statistics of core VLN training/augmentation corpora (top) and our skill-specific synthetic datasets (bottom).}

    \label{tab:dataset_statistics}
    
    \resizebox{0.9\linewidth}{!}{

    \begin{tabular}{lrrr}
\toprule
Dataset       & $\#$ Instr & $\#$ Vocab & Instr Len \\
\midrule
R2R          & $14,039$    & $4,597$     & $26.28$   \\
R2R~(ScaleVLN-Aug) & $2,891,134$ & $222$ & $25.90$ \\ 
R2R~(SRDF-Aug) & $413,999$ & $2,201$ & $23.67$ \\
GSA-R2R      & $4,675$     & $2,797$     & $26.06$   \\


\midrule
Temporal      & $2,000$     & $1,653$     & $56.60$   \\

Direction     & $450$      & $707$      & $26.78$   \\
Vertical      & $450$      & $705$      & $26.23$   \\
Stop          & $450$      & $774$      & $27.03$   \\
Landmark      & $450$      & $1,025$     & $27.62$   \\
Region        & $450$      & $971$      & $27.50$   \\
\bottomrule
\end{tabular}
    }    
\end{table}

We build six synthetic datasets where each trajectory–instruction pair targets a single skill. Paths are sampled from random Matterport3D~\citep{Matterport3D} nodes, filtered with skill-specific heuristics (\eg, frequent turns for Direction Adjustment), and kept short (4–7 steps) for comparable difficulty; Appendix~\ref{sec:data_synthesis_appendix} details the filters and Figure~\ref{fig:path_length_distribution} in Appendix~\ref{sec:trajectory_generation_appendix} shows path length analysis. 
%
%
For instruction generation, we feed each trajectory’s observations to GPT-4o~\citep{openai2024gpt4o} with skill cues so the text stays fluent and R2R-like while emphasizing the target skill (Table~\ref{tab:dataset_statistics}); Appendix~\ref{sec:synthetic_data_bias_learning} reports bias checks. 
GPT-4o is used only to synthesize diverse training data for the skill agents, while the action router operates zero-shot on benchmark instructions, reducing the risk of generator-router stylistic self-alignment.

Training uses two stages and yields two SkillNav variants. 
Stage 1 fine-tunes DUET on R2R plus ScaleVLN augmentation and the Temporal synthetic set (or SRDF augmentation for the SRDF variant) to get a skill-agnostic backbone; Appendix~\ref{sec:temporal_order_planning_agent} analyzes the Temporal Order Planning agent. 
Stage 2 fine-tunes that backbone on each skill dataset to produce five specialists: Direction Adjustment ($\pi_{\text{da}}$), Vertical Movement ($\pi_{\text{vm}}$), Stop and Pause ($\pi_{\text{sp}}$), Landmark Detection ($\pi_{\text{ld}}$), and Area/Region Identification ($\pi_{\text{ar}}$), denoted $\mathcal{S} = \{\pi_{\text{da}}, \pi_{\text{vm}}, \pi_{\text{sp}}, \pi_{\text{ld}}, \pi_{\text{ar}}\}$. Unless stated, SkillNav refers to the ScaleVLN-Aug variant.

\begin{table*}[t]
    \centering


    \caption{Performance comparison on R2R and GSA-R2R benchmarks. $^\dagger$ indicates large-scale data augmentation; $-$ indicates the unavailable statistics; bold is best, underline is second-best, and $\Delta$ rows report SkillNav minus its base (ScaleVLN/SRDF), with dark gray for gains and light gray for drops. 
    Supervised and skill-based models are trained on R2R only and without GSA-R2R fine-tuning. 
    SkillNav variants follow the ScaleVLN and SRDF settings with their respective augmentations.
    }

    \label{tab:model_performance}
    \resizebox{\linewidth}{!}{
    
    \begin{tabular}{l | c | cccc | cccc | cc | cc | cc}
    \toprule
    \multirow{3}{*}{\textbf{Methods}} 
    & \multirow{3}{*}{\#} 
    & \multicolumn{8}{c|}{\textbf{R2R}} 
    & \multicolumn{6}{c}{\textbf{GSA-R2R}} \\
    \cmidrule(lr){3-10} \cmidrule(lr){11-16}
    & 
    & \multicolumn{4}{c|}{\textbf{Val-Unseen}} 
    & \multicolumn{4}{c|}{\textbf{Test-Unseen}} 
    & \multicolumn{2}{c|}{\textbf{Test-R-Basic}} 
    & \multicolumn{2}{c}{\textbf{Test-N-Basic}}
    & \multicolumn{2}{c}{\textbf{Test-N-Scene}}\\
    & 
    & NE$\downarrow$ & OSR$\uparrow$ & SR$\uparrow$ & SPL$\uparrow$ 
    & NE$\downarrow$ & OSR$\uparrow$ & SR$\uparrow$ & SPL$\uparrow$ 
    & SR$\uparrow$ & SPL$\uparrow$ & SR$\uparrow$ & SPL$\uparrow$ 
    & SR$\uparrow$ & SPL$\uparrow$ \\
    
    \midrule
    
    \rowcolor{gray!20}
    \multicolumn{16}{c}{\textit{LLM-based VLN}} \\
    MapGPT~(GPT4v)~\citep{chen_mapgpt_2024}& 1 & $5.63$ & $58$ & $44$ & $35$ & -- & -- & -- & -- & $34$ & $30$ & $25$ & $23$ & $25$ & $23$ \\
    NavCoT~(LLaMA2)~\citep{lin_navcot_2024} & 2 & $6.26$ & $42$ & $34$ & $29$ & -- & -- & -- & -- & $37$ & $35$ & $29$ & $26$ & $29$ & $26$ \\
    NavGPT-2~(FlanT5-5B)~\citep{zhou_navgpt-2_2024} & 3 & $3.13$ & $81$ & $72$ & $61$ & $3.33$ & $80$ & $72$ & $60$ & $58$ & $45$ & $48$ & $35$ & $\mathbf{57}$ & $43$  \\
 
    NaviLLM~(Vicuna-7B)~\citep{zheng2024towards_navillm} & 4 & $3.51$ & -- & $67$ & $59$ & $3.71$ & -- & $68$ & $60$ & -- & -- & -- & -- & -- & -- \\
    DiscussNav~(GPT4)~\citep{long2024discuss} & 5 & $5.32$ & $61$ & $43$ & $40$ & -- & -- & -- & --   & -- & -- & -- & -- & -- & -- \\ 
    
    \midrule

    \rowcolor{gray!20}
    \multicolumn{16}{c}{\textit{Supervised VLN}} \\
    HAMT~\citep{chen2021history} & 6 & $2.29$ & -- & $66$ & $61$ & $3.93$ & $72$ & $65$ & $60$ & $48$ & $44$ & $42$ & $38$ & $34$ & $30$ \\
    DUET~\citep{chen_think_2022}& 7 & $3.31$ & $81$ & $72$ & $60$ & $3.65$ & $76$ & $69$ & $59$ & $58$ & $47$ & $48$ & $37$ & $40$ & $30$ \\
    BEVBERT~\citep{an2023bevbert} & 8 & $2.81$ & $84$ & $75$ & $64$ & $3.13$ & $81$ & $73$ & $62$ & $58$ & $45$ & $46$ & $35$ & $39$ & $27$  \\
    GR-DUET~\citep{hong_general_2025} & 9 & -- & -- & -- & -- & -- & -- & -- & -- & $69$ & $64$ & $57$ & $52$ & $48$ & $43$\\

    SAME~\citep{zhou2024same}$^\dagger$ & 10 & $2.73$ & -- & $76$ & $66$ & $3.03$ & -- & $74$ & $64$ & -- & -- & -- & -- & -- & -- \\
    
    ScaleVLN~\citep{wang2023scaling} $^\dagger$ & 11 & $2.39$ & {$88$} & $79$ & $70$ & $2.73$ & $84$ & $77$ & $68$ & \underline{$78$} & \underline{$67$} & \underline{$69$} & \underline{$57$} & \underline{$55$} & $43$\\


    SRDF~\citep{wang_bootstrapping_2024} $^\dagger$ & 12 & \underline{$1.83$} & $\underline{89}$ & $\underline{84}$ & $\underline{78}$ & \underline{$1.88$} & $\mathbf{88}$ & $\mathbf{84}$ & $\underline{77}$ & $71$ & $63$ & $59$ & $49$ & $52$ & $43$ \\


    \midrule
    
    \rowcolor{gray!20}
    \multicolumn{16}{c}{\textit{Mixture of Skill-based VLN}} \\


    SkillNav~(ScaleVLN-Aug)$^\dagger$ & 13 & $1.97$ & $\mathbf{89}$ & {$83$} & {$77$} & $2.53$ & $83$ & $78$ & \underline{$70$} & $\mathbf{79}$ & $\mathbf{69}$ & $\mathbf{72}$ & $\mathbf{61}$ & $\mathbf{57}$ & $\mathbf{48}$ \\

        

    \quad  $\Delta$ SkillNav~(ScaleVLN-Aug) - ScaleVLN &   & \gn{$-0.42$} & \gn{$+1.77$} & \gn{$+3.36$} & \gn{$+6.54$} & \gn{$-0.20$} & \dpn{$-1.65$} & \gn{$+0.88$} & \gn{$+1.80$} & \gn{$+0.71$} & \gn{$+2.18$} & \gn{$+2.45$} & \gn{$+4.18$} & \gn{$+2.16$} & \gn{$+5.26$} \\


    SkillNav~(SRDF-Aug)$^\dagger$& 14 & $\mathbf{1.79}$ & ${89}$ & $\mathbf{84}$ & $\mathbf{78}$ & $\mathbf{1.76}$ & \underline{$87$} & $\underline{84}$ & $\mathbf{77}$ & $71$ & $64$ & $61$ & $50$ & $54$ & \underline{$45$}  \\

    

    \quad  $\Delta$ SkillNav~(SRDF-Aug) - SRDF &   & \gn{$-0.04$} & \dpn{$-0.26$} & \gn{$+0.20$} & \gn{$+0.22$} & \gn{$-0.12$} & \dpn{$-0.82$} & \dpn{$-0.28$} & \gn{$+0.09$} & \gn{$+0.56$} & \gn{$+1.02$} & \gn{$+2.83$} & \gn{$+0.87$} & \gn{$+2.78$} & \gn{$+2.02$} \\


        


    


    \bottomrule
    \end{tabular}
    }
\end{table*}


\subsection{VLM-based Action Router}\label{sec:action_router}


After training specialized agents for distinct navigation skills, we integrate them into the SkillNav framework using a VLM-based Action Router to coordinate their execution. 
Inspired by planning systems such as LLM-Planner~\citep{song2023llmplanner}, Mic~\citep{qiao2023march}, and A2Nav~\citep{chen2023a2nav}, our router leverages a large VLM (\eg, GPT-4o~\citep{openai2024gpt4o}, Qwen2.5-VL-7B-Instruct~\citep{Qwen2.5-VL}) in a zero-shot in-context fashion to dynamically select the most suitable agent at each time step. 
%
%
We structure the routing process into three distinct reasoning phases:

\noindent\textbf{Phase 1: Temporal Reordering Module.}\label{sec:temporal_reordering}
The Temporal Reordering Module only takes the original natural language instruction as input. It applies the instruction reordering prompt to turn navigation instructions into 
a list of 
subgoals $I_{\text{reorder}}$.
It follows the four temporal relations 
described in the Temporal Order Planning skill in Section~\ref{sec:skill_taxonomy}, 
making implicit temporal details explicit and ensuring the correct subgoal execution order. This procedure is formulated as
\begin{equation}
    I_{\text{reorder}} = \text{LLM}_{\text{TemporalReorder}}(I).
\end{equation}

\noindent\textbf{Phase 2: Subgoal Localizer.} Given the reordered subgoals $ I_{\text{reorder}} =  [p_1, p_2, \ldots, p_m]$, observed history $H_{t-1}$,
and the sequence of previously executed subgoals \(G_{t-1} = [p^*_1, \dots, p^*_{t-1} ]\), the model identifies the next subgoal \(p^*_{t}\) to be executed for the current time step \(t\) and outputs the corresponding reasoning trace \( r_t \), later used by the router for decision verification.
The output can be formalized as:
\begin{equation}
    p^*_t, r_t = \texttt{Localize}(I_{\text{reorder}}, H_{t-1} , G_{t-1}).
\end{equation}
The sequence of executed subgoals is then updated as:
\begin{equation}
     G_{t} =  G_{t-1} \parallel p^*_t.
\end{equation}
\noindent\textbf{Phase 3: Skill Router.}
At time step $t$, the skill router determines which skill-based agent \( \pi^*_t \in \mathcal{S} \) is most appropriate for executing the selected subgoal \(p_t^{*} \).
Additionally, it receives the original instruction \( I \) as a part of the input context 
to capture additional linguistic cues such as verbs and spatial references. It also uses the reasoning trace \( r_t \) from Phase 2 to enhance its understanding of the current subgoal. At each step, exactly one skill is selected, formulated as
\begin{equation}
    \pi^*_t = \arg\max_{\pi \in \mathcal{S}} \texttt{Router}(I, p^*_{t}, r_{t}).
\end{equation}
Once the appropriate skill-based agent is selected, 
it is invoked by the following Eq.~\eqref{eq:duet} to predict the navigation action at time step $t$:
 \begin{equation}
    a_{t}^{*} = \pi^{*}_{t}(I, O_t, M_t).
 \end{equation}
%
%
Our proposedrouter enables modular skill execution by integrating natural language, visual inputs, and observed history, using the Temporal Reordering LLM to effectively bridge instructions with actionable skill modules.

\section{Experiments}

\subsection{Experiment Setup}
\noindent\textbf{Evaluation Datasets.} 
We primarily use the Room-to-Room (R2R) dataset~\citep{anderson2018vision}, especially the 
unseen split of validation (Val Unseen) and test (Test Unseen) splits. 
R2R is a commonly used benchmark in VLN consisting of panoramic RGB-D scans from the Matterport3D~\citep{Matterport3D} simulator and providing crowd-sourced instructions paired with navigation paths. 
Moreover, we evaluate the generalization ability of SkillNav on GSA-R2R~\citep{hong_general_2025}, which includes residential (R) and non-residential (N) scenes (\textit{e.g.}, shops, restaurants, and museums) from Habitat-Matterport3D~\citep{ramakrishnan2021hm3d}, and diverse instruction styles with role-specific dialogues (\textit{e.g.}, travel guides (Scene) beyond the basic style of R2R (Basic)).
To demonstrate SkillNav's robustness in long-horizon, fine-grained instruction scenarios, we conduct additional experiments
on RxR-English~\cite{ku_room-across-room_2020, wang_bootstrapping_2024}, which is more linguistically detailed and has significantly longer trajectory paths
than R2R.

\noindent\textbf{Evaluation Metrics.}
We use the standard metrics to evaluate the navigation performance~\citep{anderson2018vision, zhao_mind_2023, qi_reverie_2020}: (1) Navigation Error (NE): the distance between the stop location and the target; (2) Oracle Success Rate (OSR): the agent ever gets close enough to the goal at any point along its trajectory, regardless of where it decides to stop; (3) Success Rate (SR): the ratio of agents stopping within $3$ meters of the target; 
(4) Success rate weighted by Path Length (SPL): measures navigation efficiency by weighting the success rate with the ratio between the shortest path length and the agent’s actual path length, penalizing unnecessarily long trajectories; (5) Normalized Dynamic Time
Warping (nDTW): penalizes deviations from the ground-truth paths.


\noindent\textbf{Implementation Details.}\label{sec:implementation}
%
We develop two SkillNav variants tailored to the ScaleVLN and SRDF settings, respectively. 
For the ScaleVLN configuration, we employ CLIP-B/16~\citep{radford2021learning} as the visual encoder and BERT-base-uncased~\citep{bert-base-uncased} as the language backbone. Conversely, the SRDF setup swaps the visual encoder to InternViT-6B~\citep{chen2024internvl}.
In our SkillNav VLM-based action router, we adopt GPT-4o~\citep{openai2024gpt4o} as the temporal reordering module due to its superior instruction-following capabilities and employ Qwen2.5-VL-7B-Instruct~\citep{Qwen2.5-VL} for the other two phases because of its strong multi-modal alignment and reasoning abilities. All inferences with the action router are performed using in-context prompting. 
Unless otherwise specified, SkillNav denotes the ScaleVLN setting variant by default.
Additional skill-based agent training, hyperparameter, and routing details are provided in Appendix~\ref{sec:appendix_reproducibility}.  
%

\begin{figure}[!t]
    \centering
    \includegraphics[width=\linewidth]{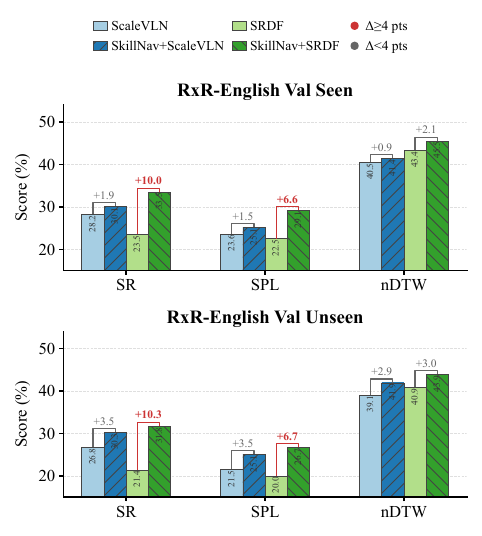}
    \caption{Zero-shot generalization results on RxR-English. SkillNav improves both ScaleVLN and SRDF backbones across SR, SPL, and nDTW on Val Seen and Val Unseen splits.}
    \label{fig:rxr_generalization}
\end{figure}

\begin{figure*}[!t]
    \centering
    \includegraphics[width=\linewidth]{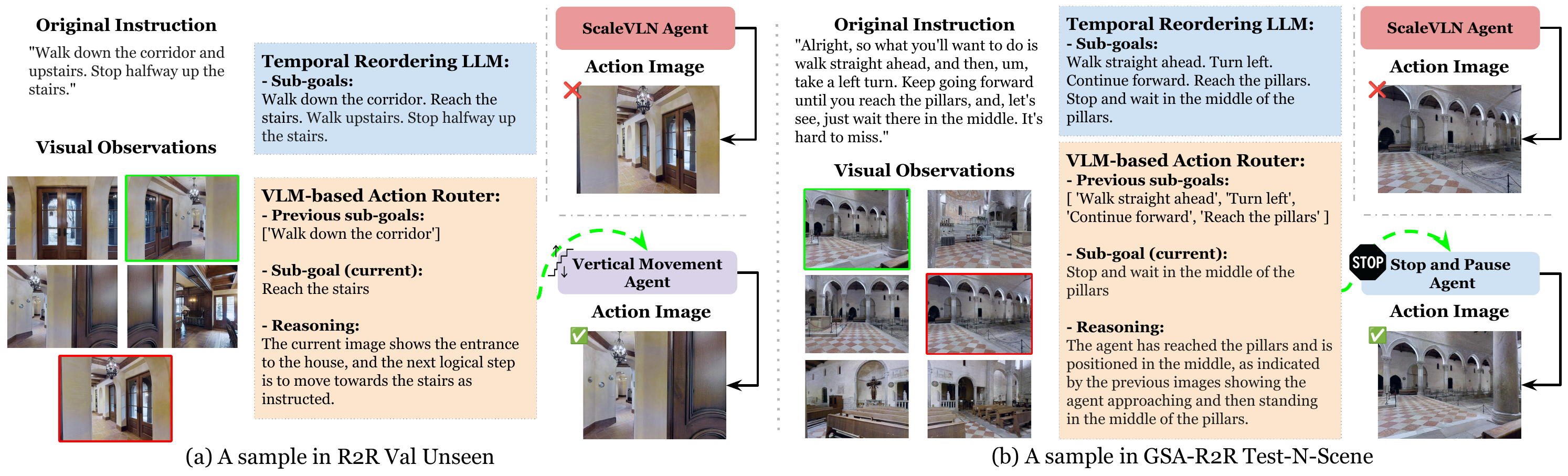}
    \caption{Qualitative examples of routing and navigation results. These examples include cases where the instruction is temporally complex, colloquial, or spatially ambiguous. 
    }
    \label{fig:qualitative_example}
\end{figure*}

\begin{table}[t]
\centering
\caption{Temporal reordering ablation for the VLM-based action router on GSA-R2R across residential (R) and non-residential (N) scenarios with Basic/Scene instructions.
Reorder: \ding{55} disables the LLM-guided Temporal Reordering module; \ding{52} enables it, and the ordered subgoal plan is fed only to the router while skill agents still receive the original instruction. Router: includes the Subgoal Localizer and Skill Router phases; 
\textbf{Qwen}: Qwen2.5-VL-7B-Instruct; \textbf{GLM}: GLM-4.1V-9B-Thinking.
}
\label{tab:temporal_reorder_ablation}
\resizebox{\linewidth}{!}{
\begin{tabular}{cc|cc|cc|cc}
\toprule
\multirow{2}{*}{\textbf{Reorder}} & \multirow{2}{*}{\textbf{Router}} 
& \multicolumn{2}{c|}{\textbf{Test-R-Basic}} 
& \multicolumn{2}{c|}{\textbf{Test-N-Basic}} 
& \multicolumn{2}{c}{\textbf{Test-N-Scene}} \\
\cmidrule(lr){3-4} \cmidrule(lr){5-6} \cmidrule(lr){7-8}
 &  & \textbf{SR} & \textbf{SPL} & \textbf{SR} & \textbf{SPL} & \textbf{SR} & \textbf{SPL} \\
\midrule
\ding{55} & Qwen & $78.42$ & $67.80$ & $71.01$ & $59.62$ & $55.46$ & $45.43$ \\
\ding{52} & Qwen & $\mathbf{78.83}$ & $\mathbf{68.88}$ & $\mathbf{71.58}$ & $\mathbf{61.34}$ & $56.66$ & $\mathbf{47.96}$ \\

\ding{55} & GLM & $77.46$ & $66.27$ & $70.70$ & $58.63$ & $55.62$ & $42.64$ \\
\ding{52} & GLM & $78.60$ & $67.93$ & $71.13$ & $59.73$ & $\mathbf{56.80}$ & $46.51$ \\
\bottomrule
\end{tabular}
}
\end{table}

\begin{table}[t]
\centering
\caption{
VLM choice for subgoal localization and skill routing on GSA-R2R.
The Temporal Reordering module is fixed to \textit{enabled}.
Router Definitions: \textbf{Random} selects skill-based agents without action routing; 
\textbf{GPT-4o} (GPT-4o-2024-08-06); 
\textbf{Qwen} and \textbf{GLM} follow the definitions in Table~\ref{tab:temporal_reorder_ablation}.
Results indicate that once temporal structure is provided, performance differences across routers become small, indicating robustness to the specific VLM router.}
\label{tab:exp_router_vlm_ablation}
\resizebox{\linewidth}{!}{
\begin{tabular}{l|cc|cc|cc}
\toprule
\multirow{2}{*}{\textbf{Router}} 
& \multicolumn{2}{c|}{\textbf{Test-R-Basic}} 
& \multicolumn{2}{c|}{\textbf{Test-N-Basic}} 
& \multicolumn{2}{c}{\textbf{Test-N-Scene}} \\
\cmidrule(lr){2-3} \cmidrule(lr){4-5} \cmidrule(lr){6-7}
 & \textbf{SR} & \textbf{SPL} & \textbf{SR} & \textbf{SPL} & \textbf{SR} & \textbf{SPL} \\
\midrule
Random & $78.39$ & $67.46$ & $70.93$ & $59.71$ & $54.61$ & $43.17$ \\
GLM    & $78.60$ & $67.93$ & $71.13$ & $59.73$ & \underline{$56.80$} & $46.51$ \\
Qwen   & \underline{$78.83$} & \underline{$68.88$} & \underline{$71.58$} & \underline{$61.34$} & $56.66$ & \underline{$47.96$} \\
GPT-4o & $\mathbf{79.41}$ & $\mathbf{69.18}$ & $\mathbf{72.75}$ & $\mathbf{62.48}$ & $\mathbf{58.16}$ & $\mathbf{48.96}$ \\
\bottomrule
\end{tabular}
}
\end{table}

\subsection{Main Results}




%
Table~\ref{tab:model_performance} compares SkillNav against state-of-the-art VLN agents, demonstrating its superior generalization to novel instructions and environments while maintaining competitive performance on the R2R benchmark. 
Regarding generalization, while the strong baseline SRDF (Method~\#12) overfits to R2R's specific style, the ScaleVLN-based SkillNav (Method~\#13) excels, outperforming baselines on the challenging GSA-R2R Test-N-Scene ($48\%$ vs. $43\%$ SPL). 
%
Moreover, in Figure~\ref{fig:rxr_generalization}, SkillNav also achieves superior generalization with absolute nDTW gains of up to +3.04\% on RxR-English Val Unseen, confirming that our modular decomposition successfully helps the agent adhere to dense, descriptive instructions better than monolithic baselines.
Regarding R2R, SkillNav consistently improves or matches its specific backbones: the ScaleVLN-based variant boosts its baseline (Method~\#11) by $\sim6.5\%$ SPL on Val-Unseen, while the SRDF-based variant (Method~\#14) maintains parity with the state-of-the-art SRDF at $77\%$ SPL, successfully balancing versatility with precision.
However, current agents are effectively matching human-level performance on R2R, indicating that the remaining headroom for large absolute gains on R2R is limited (see Table~\ref{tab:r2r_human_upper_bound} in Appendix~\ref{sec:r2r_gsa_gap_appendix}). 
\subsection{Ablation Study and Analysis}\label{sec:ablation_study}

\begin{table*}[!t]
    \centering 
    \caption{Evaluation of each skill-based agent on the NavNuances benchmark across four skill categories: Direction Change (DC), Vertical Movement (VM), Landmark Recognition (LR), and Region Recognition (RR). 
    Following the NavNuances, evaluation metrics differ across skill subsets: DC and LR are reported only with SR, VM includes SR/OSR/SPL, and RR provides SR/OSR. We retain this heterogeneous metric design to ensure comparability with prior work.
    Ident.: Identification.
    }
    \label{tab:performance_navnuance}
    \small
    \setlength{\tabcolsep}{3.5pt}
    \resizebox{0.9\textwidth}{!}{
    \begin{tabular}{ll|c|ccc|c|cc}
        \toprule
         & \multirow{2}{*}{\textbf{Methods}} 
        & \multicolumn{1}{c|}{DC} 
        & \multicolumn{3}{c|}{VM} 
        & \multicolumn{1}{c|}{LR} 
        & \multicolumn{2}{c}{RR} \\
        \cmidrule(lr){3-3}  \cmidrule(lr){4-6} \cmidrule(lr){7-7} \cmidrule(lr){8-9} 
    
         & & SR & SR & OSR & SPL  & SR & SR & OSR \\
        \midrule
        \multirow{2}{*}{\textbf{VLN Agents}}  & ScaleVLN~\citep{wang2023scaling} & $68.39$ & $81.76$ & $88.82$ & $76.34$ & $28.32$ & $82.91$ & $95.27$ \\
    
        & SRDF~\citep{wang_bootstrapping_2024} & $59.93$ & $82.94$ & $\mathbf{91.18}$ & $80.98$ & $26.28$ & $77.09$ & $94.55$ \\

        & Mixed Skills & $66.84$ & $84.11$ & $87.65$ & $79.22$ & $\mathbf{48.90}$ & $81.82$ & $90.91$ \\ 
        \midrule
    

      
    
        \multirow{5}{*}{\textbf{Skill-based Agents}} & Direction Adjustment   & $\mathbf{70.81}$ & $81.76$ & $\mathbf{91.18}$ & $76.28$ & $31.39$ & $81.82$ & $94.91$ \\
    
        & Vertical Movement  & $70.68$ & $\mathbf{87.65}$ & $89.41$ & $\mathbf{83.83}$ & $30.22$ & $82.18$ & $96.00$  \\
    
        & Landmark Detection & $70.29$ & $82.35$ & $85.29$ & $78.94$ & $\mathbf{31.53}$ & $83.64$ & $\mathbf{97.09}$\\
    
        & Area and Region Ident. & $67.53$ & $84.12$ & $88.82$ & $80.49$ & $29.20$ & $\mathbf{85.09}$ & $96.36$ \\

        & Stop and Pause & $68.91$ & $84.71$ & $87.06$ & $80.67$ & $29.78$ & $83.64$ & $\mathbf{97.09}$ \\
        
        \bottomrule
        \end{tabular}
        }

\end{table*}

\begin{table}[h]
\centering
\small
\caption{Expert call distribution on GSA-R2R Test-N-Scene. The values indicate the percentage of steps each specialist policy ($\pi$) was activated by the router.}
\label{tab:expert_call_distribution}
\resizebox{\linewidth}{!}{
\begin{tabular}{ccccc}
\toprule
\textbf{$\pi_{\text{sp}}$} & \textbf{$\pi_{\text{da}}$} & \textbf{$\pi_{\text{ar}}$} & \textbf{$\pi_{\text{ld}}$} & \textbf{$\pi_{\text{vm}}$} \\
\midrule
$34.42\%$ & $23.61\%$ & $18.75\%$ & $14.23\%$ & $8.99\%$ \\
\bottomrule
\end{tabular}
}
\end{table}

\noindent\textbf{Action Router.}
We evaluate the Action Router structural mechanism in Table~\ref{tab:temporal_reorder_ablation} and backbone capability in Table~\ref{tab:exp_router_vlm_ablation}. We maintain that operating on raw instructions forces difficult implicit temporal reasoning that limits open-world generalization~\citep{physicalintelligence2025pi05}. Disabling Temporal Reordering causes consistent performance drops, marked by a $2.5\%$ SPL decline on Test-N-Scene, which confirms that explicit decomposition acts as a necessary structural scaffold. Complementing this, Table~\ref{tab:exp_router_vlm_ablation} shows that while a Random baseline suffices for basic instructions, novel environments require advanced reasoning. GPT-4o demonstrates the strongest generalization, proving that explicit reasoning and structural priors are key differentiators for handling unseen scenes.

\noindent\textbf{Skill Evaluation.}
We validate our skill-based agents on NavNuances, where categorization into four atomic skills reveals that each agent excels in its corresponding domain in Table~\ref{tab:performance_navnuance}. These results confirm that targeted supervision fosters functional specialization superior to SOTA VLN baselines in isolated skill settings. See Appendix~\ref{sec:appendix_skill_mixed} for mixed-skill analysis.
Furthermore, Table~\ref{tab:expert_call_distribution} reveals that the agent adopts a conservative, precision-first strategy. The dominance of the control-focused policies defined in Section~\ref{sec:data_synthesis_agent_train}, specifically $\pi_{\text{sp}}$ and $\pi_{\text{da}}$ (nearly 60\%), indicates that robust navigation depends less on constant semantic recognition and more on continuous state verification. In contrast, semantic skills ($\pi_{\text{ar}}$, $\pi_{\text{ld}}$) act as sparse anchors, activated only when specific visual grounding is strictly necessary for decision-making.
In addition, we provide a detailed skill routing analysis in Appendix~\ref{sec:appendix_skill_routing}.

\noindent \textbf{Error Analysis.}
We conduct a manual error analysis of 17 failure cases (detailed in Appendix~\ref{sec:appendix_human_eval_skill_routing}). Our findings indicate that the primary bottleneck is not the router's linguistic reasoning, but rather visual grounding deficits and textual-prior biases. While the router typically parses the correct linguistic subgoal (e.g., ``stop at the sink''), it frequently struggles to bind these textual entities to their corresponding visual referents in cluttered scenes. Furthermore, we observe a systematic textual-prior bias: the router underutilizes fine-grained Landmark Detection, erroneously defaulting to coarser Region Identification or Direction skills even when object-level grounding is required. Beyond these core issues, we categorize three supplementary failure modalities: (1) Initial Misalignment, where incorrect starting orientations cause cumulative trajectory drift; (2) Subgoal Hallucination, wherein the VLM router generates spurious subgoals ungrounded in the physical environment; and (3) Controller Bottlenecks, where execution fails due to dense, ``stacked'' commands that exceed the capacity of the underlying expert policies.

\subsection{Efficiency Analysis}\label{sec:efficiency_analysis}
\noindent\textbf{Training Cost.}
Fine-tuning five skills on the Temporal Order Planning agent with R2R and synthetic skill-specific datasets under the SRDF setting requires approximately $3,329$ minutes ($\sim 55.5$ hours) in total.
For comparison, SRDF training on R2R with large data augmentation takes $2,521$ minutes ($\sim 42$ hours), suggesting that SkillNav's skill-based training introduces a relatively higher training cost. 
However, this represents a one-time training investment; unlike prior supervised VLN models that require repeated retraining to adapt to new environments or instruction styles, SkillNav achieves strong generalization across datasets without additional retraining.
\noindent\textbf{Inference Cost.}
We provide inference time and throughput comparison in Table~\ref{tab:runtime}.
SkillNav introduces overhead due to the VLM-based action router, reaching $0.49$ throughput on Test-N-Basic of GSA-R2R, which is roughly $50\times $ slower than ScaleVLN but still nearly $20\times$ faster than MapGPT.
%
%
%
%
We further analyze the per-step latency of open-source VLM models in Table~\ref{tab:router_runtime_analysis_appendix} and the computation cost of closed-source models in Table~\ref{tab:efficiency_routers_deployment_costs}. Local open-source models such as Qwen2.5-VL-7B achieve a  competitive inference time of $1.57$s per step.
By decoupling high-level reasoning from low-level action execution, the VLM-based Router is invoked only during skill-switching events rather than at every execution step. This allows SkillNav to achieve a latency of $9.69$s per case, which is $2-4\times$ faster than existing baselines like NavGPT~\cite{zhou_navgpt_2023} and
FlexVLN~\cite{zhang2025flexvln}.


Overall, while SkillNav is less efficient than supervised models, it achieves a better efficiency-generalization trade-off. Also, it advances both efficiency and generalization compared to LLM-based VLN agents.

\subsection{Qualitative Examples}
Figure~\ref{fig:qualitative_example} highlights SkillNav's dynamic decision-making capabilities, showing how the framework effectively reorders temporal instruction plans and identifies active subgoals to select the precise skill and action required for every navigation step.
Specifically, in example (a), the router correctly identifies the destination (pillars) to trigger the stop action, while in (b), it detects the need for elevation and selects the vertical movement skill. 
These cases confirm the framework's robustness across diverse instruction styles and environments.

\begin{table}[t]
\centering
\caption{Runtime and throughput of baselines and SkillNav. 
Numbers are wall-clock runtime in seconds. 
Random = randomly select skill-based agents without utilizing the action router.
}
\label{tab:runtime}
\resizebox{\linewidth}{!}{
\begin{tabular}{l l r r}
\toprule
\textbf{Method} & \textbf{Split} & \textbf{Runtime (s)} & \textbf{Inferences/s} \\
\midrule

\rowcolor{gray!20}
\multicolumn{4}{c}{\textit{Supervised VLN}} \\

\multirow{2}{*}{ScaleVLN} & Test-R-Basic & $513.8$   & $\mathbf{28.03}$ \\
& Test-N-Basic & $342.7$   & $\mathbf{26.26}$ \\
\midrule

\rowcolor{gray!20}
\multicolumn{4}{c}{\textit{LLM-based VLN}} \\

\multirow{2}{*}{MapGPT} & Test-R-Basic & $\sim597,000$ & $0.02$ \\
& Test-N-Basic & $\sim373,000$ & $0.02$ \\
\midrule

\rowcolor{gray!20}
\multicolumn{4}{c}{\textit{Our Mixture of Skill-based VLN}} \\


\multirow{2}{*}{SkillNav} & Test-R-Basic & $\sim27,000$  & $0.54$ \\
& Test-N-Basic & $\sim18,360$  & $0.49$ \\
\bottomrule
\end{tabular}
}
\end{table}

\section{Conclusion}

We introduce SkillNav, a VLN framework combining skill-based learning with VLM-based routing to dynamically select expert policies. 
By decoupling high-level decision-making from low-level execution, SkillNav leverages an LLM-based temporal reordering module to break down intricate instructions into structured sequences, while simultaneously decomposing the navigation task into atomic, manageable skills.
Evaluations demonstrate strong performance on R2R and superior generalization on GSA-R2R. While incurring moderate overhead, SkillNav achieves a better efficiency-performance trade-off than pure VLM or supervised baselines. 
This interpretable approach advances compositional reasoning and robustness in VLN.
Our framework provides a novel and interpretable approach that advances compositional reasoning and generalization for the VLN research community.

\section*{Limitations}\label{sec:limitations}

SkillNav is evaluated in discrete VLN simulator environments, including R2R, GSA-R2R, and NavNuances. This setting is standard for isolating language grounding, semantic planning, and navigation decision making, but it does not fully test continuous control or real-world robotic deployment. The proposed router and semantic skill taxonomy operate over language, visual context, and subgoals, so they are not tied to graph-specific transitions. However, each low-level skill executor would need adaptation when the action space changes from selecting neighboring viewpoints to continuous velocity, rotation, waypoint, or collision-avoidance control. Extending SkillNav to VLN-CE, Habitat continuous navigation, and physical robots is therefore an important next step.

The current skill library is also not intended to be universally exhaustive. The selected skills cover frequent indoor VLN phenomena. More specialized domains may require additional skill executors, such as object manipulation, transparent material handling, or human-aware navigation. SkillNav is designed to support this extension by adding new skill-specific data and a new executor while keeping the high-level router structure unchanged.

Finally, SkillNav introduces additional inference cost because it uses LLM/VLM reasoning for temporal reordering and action routing. Our efficiency analysis shows that this cost is lower than LLM-heavy VLN agents but higher than monolithic supervised agents. This trade-off may limit deployment in latency-constrained settings unless the router is distilled, cached, or replaced by a smaller task-specific model.

\section{Acknowledgment}

This project is partially supported by the Office of Naval Research (ONR) grant N00014-23-1-2417. Any opinions, findings, conclusions, or recommendations expressed in this material are those of the authors and do not necessarily reflect the views of the Office of Naval Research.
Zehao Wang is supported by KULeuven Methusalem project Lifelines.

\bibliography{custom}

\begin{thebibliography}{47}
\providecommand{\natexlab}[1]{#1}

\bibitem[{An et~al.(2023)An, Qi, Li, Huang, Wang, Tan, and
  Shao}]{an2023bevbert}
Dong An, Yuankai Qi, Yangguang Li, Yan Huang, Liang Wang, Tieniu Tan, and Jing
  Shao. 2023.
\newblock Bevbert: Multimodal map pre-training for language-guided navigation.
\newblock \emph{Proceedings of the IEEE/CVF International Conference on
  Computer Vision}.

\bibitem[{Anderson et~al.(2018)Anderson, Wu, Teney, Bruce, Johnson,
  S{\"u}nderhauf, Reid, Gould, and Van Den~Hengel}]{anderson2018vision}
Peter Anderson, Qi~Wu, Damien Teney, Jake Bruce, Mark Johnson, Niko
  S{\"u}nderhauf, Ian Reid, Stephen Gould, and Anton Van Den~Hengel. 2018.
\newblock Vision-and-language navigation: Interpreting visually-grounded
  navigation instructions in real environments.
\newblock In \emph{Proceedings of the IEEE conference on computer vision and
  pattern recognition}, pages 3674--3683.

\bibitem[{Bai et~al.(2025)Bai, Chen, Liu, Wang, Ge, Song, Dang, Wang, Wang,
  Tang, Zhong, Zhu, Yang, Li, Wan, Wang, Ding, Fu, Xu, Ye, Zhang, Xie, Cheng,
  Zhang, Yang, Xu, and Lin}]{Qwen2.5-VL}
Shuai Bai, Keqin Chen, Xuejing Liu, Jialin Wang, Wenbin Ge, Sibo Song, Kai
  Dang, Peng Wang, Shijie Wang, Jun Tang, Humen Zhong, Yuanzhi Zhu, Mingkun
  Yang, Zhaohai Li, Jianqiang Wan, Pengfei Wang, Wei Ding, Zheren Fu, Yiheng
  Xu, and 8 others. 2025.
\newblock Qwen2.5-vl technical report.
\newblock \emph{arXiv preprint arXiv:2502.13923}.

\bibitem[{Chang et~al.(2017)Chang, Dai, Funkhouser, Halber, Niessner, Savva,
  Song, Zeng, and Zhang}]{Matterport3D}
Angel Chang, Angela Dai, Thomas Funkhouser, Maciej Halber, Matthias Niessner,
  Manolis Savva, Shuran Song, Andy Zeng, and Yinda Zhang. 2017.
\newblock Matterport3d: Learning from rgb-d data in indoor environments.
\newblock \emph{International Conference on 3D Vision (3DV)}.

\bibitem[{Chen et~al.(2024{\natexlab{a}})Chen, Lin, Xu, Chai, Liang, and
  Wong}]{chen_mapgpt_2024}
Jiaqi Chen, Bingqian Lin, Ran Xu, Zhenhua Chai, Xiaodan Liang, and Kwan-Yee~K.
  Wong. 2024{\natexlab{a}}.
\newblock \href {https://doi.org/10.48550/arXiv.2401.07314} {{MapGPT}:
  {Map}-{Guided} {Prompting} with {Adaptive} {Path} {Planning} for
  {Vision}-and-{Language} {Navigation}}.
\newblock \emph{arXiv preprint}.
\newblock ArXiv:2401.07314 [cs].

\bibitem[{Chen et~al.(2023)Chen, Sun, Zhi, Zeng, Li, Liu, Tan, and
  Gan}]{chen2023a2nav}
Peihao Chen, Xinyu Sun, Hongyan Zhi, Runhao Zeng, Thomas~H. Li, Gaowen Liu,
  Mingkui Tan, and Chuang Gan. 2023.
\newblock \href {https://arxiv.org/abs/2308.07997} {A2nav: Action-aware
  zero-shot robot navigation by exploiting vision-and-language ability of
  foundation models}.
\newblock \emph{CoRR}, abs/2308.07997.

\bibitem[{Chen et~al.(2021)Chen, Guhur, Schmid, and Laptev}]{chen2021history}
Shizhe Chen, Pierre-Louis Guhur, Cordelia Schmid, and Ivan Laptev. 2021.
\newblock History aware multimodal transformer for vision-and-language
  navigation.
\newblock \emph{Advances in neural information processing systems},
  34:5834--5847.

\bibitem[{Chen et~al.(2022{\natexlab{a}})Chen, Guhur, Tapaswi, Schmid, and
  Laptev}]{Chen_2022_HM3D_AutoVLN}
Shizhe Chen, Pierre-Louis Guhur, Makarand Tapaswi, Cordelia Schmid, and Ivan
  Laptev. 2022{\natexlab{a}}.
\newblock Learning from unlabeled 3d environments for vision-and-language
  navigation.
\newblock In \emph{ECCV}.

\bibitem[{Chen et~al.(2022{\natexlab{b}})Chen, Guhur, Tapaswi, Schmid, and
  Laptev}]{chen_think_2022}
Shizhe Chen, Pierre-Louis Guhur, Makarand Tapaswi, Cordelia Schmid, and Ivan
  Laptev. 2022{\natexlab{b}}.
\newblock \href {https://doi.org/10.48550/arXiv.2202.11742} {Think {Global},
  {Act} {Local}: {Dual}-scale {Graph} {Transformer} for {Vision}-and-{Language}
  {Navigation}}.
\newblock \emph{arXiv preprint}.
\newblock ArXiv:2202.11742 [cs].

\bibitem[{Chen et~al.(2024{\natexlab{b}})Chen, Wu, Wang, Su, Chen, Xing, Zhong,
  Zhang, Zhu, Lu et~al.}]{chen2024internvl}
Zhe Chen, Jiannan Wu, Wenhai Wang, Weijie Su, Guo Chen, Sen Xing, Muyan Zhong,
  Qinglong Zhang, Xizhou Zhu, Lewei Lu, and 1 others. 2024{\natexlab{b}}.
\newblock Internvl: Scaling up vision foundation models and aligning for
  generic visual-linguistic tasks.
\newblock In \emph{Proceedings of the IEEE/CVF Conference on Computer Vision
  and Pattern Recognition}, pages 24185--24198.

\bibitem[{Devlin et~al.(2018)Devlin, Chang, Lee, and
  Toutanova}]{bert-base-uncased}
Jacob Devlin, Ming{-}Wei Chang, Kenton Lee, and Kristina Toutanova. 2018.
\newblock \href {https://arxiv.org/abs/1810.04805} {{BERT:} pre-training of
  deep bidirectional transformers for language understanding}.
\newblock \emph{CoRR}, abs/1810.04805.

\bibitem[{Hao et~al.(2020)Hao, Li, Li, Carin, and Gao}]{hao2020prevalent}
Weituo Hao, Chunyuan Li, Xiujun Li, Lawrence Carin, and Jianfeng Gao. 2020.
\newblock Towards learning a generic agent for vision-and-language navigation
  via pre-training.
\newblock In \emph{Proceedings of the IEEE/CVF Conference on Computer Vision
  and Pattern Recognition (CVPR)}, pages 13137--13146.

\bibitem[{Hong et~al.(2025)Hong, Qiao, Wang, Liu, and Wu}]{hong_general_2025}
Haodong Hong, Yanyuan Qiao, Sen Wang, Jiajun Liu, and Qi~Wu. 2025.
\newblock \href {https://openreview.net/forum?id=2oKkQTyfz7} {General scene
  adaptation for vision-and-language navigation}.
\newblock In \emph{The Thirteenth International Conference on Learning
  Representations}.

\bibitem[{Hong et~al.(2021)Hong, Wu, Qi, Rodriguez-Opazo, and
  Gould}]{hong_recurrent_2021}
Yicong Hong, Qi~Wu, Yuankai Qi, Cristian Rodriguez-Opazo, and Stephen Gould.
  2021.
\newblock \href {https://doi.org/10.48550/arXiv.2011.13922} {A {Recurrent}
  {Vision}-and-{Language} {BERT} for {Navigation}}.
\newblock \emph{arXiv preprint}.
\newblock ArXiv:2011.13922 [cs].

\bibitem[{Intelligence et~al.(2025)Intelligence, Black, Brown, Darpinian,
  Dhabalia, Driess, Esmail, Equi, Finn, Fusai, Galliker, Ghosh, Groom, Hausman,
  Ichter, Jakubczak, Jones, Ke, LeBlanc, Levine, Li-Bell, Mothukuri, Nair,
  Pertsch, Ren, Shi, Smith, Springenberg, Stachowicz, Tanner, Vuong, Walke,
  Walling, Wang, Yu, and Zhilinsky}]{physicalintelligence2025pi05}
Physical Intelligence, Kevin Black, Noah Brown, James Darpinian, Karan
  Dhabalia, Danny Driess, Adnan Esmail, Michael Equi, Chelsea Finn, Niccolo
  Fusai, Manuel~Y. Galliker, Dibya Ghosh, Lachy Groom, Karol Hausman, Brian
  Ichter, Szymon Jakubczak, Tim Jones, Liyiming Ke, Devin LeBlanc, and 17
  others. 2025.
\newblock $\pi_{0.5}$: a vision-language-action model with open-world
  generalization.
\newblock \emph{arXiv preprint arXiv:2504.16054}.

\bibitem[{Ku et~al.(2020)Ku, Anderson, Patel, Ie, and
  Baldridge}]{ku_room-across-room_2020}
Alexander Ku, Peter Anderson, Roma Patel, Eugene Ie, and Jason Baldridge. 2020.
\newblock \href {https://doi.org/10.48550/arXiv.2010.07954}
  {Room-{Across}-{Room}: {Multilingual} {Vision}-and-{Language} {Navigation}
  with {Dense} {Spatiotemporal} {Grounding}}.
\newblock \emph{arXiv preprint}.
\newblock ArXiv:2010.07954.

\bibitem[{Li et~al.(2024)Li, Gan, Yang, Yang, Li, Wang, Gao
  et~al.}]{li2024multimodal}
Chunyuan Li, Zhe Gan, Zhengyuan Yang, Jianwei Yang, Linjie Li, Lijuan Wang,
  Jianfeng Gao, and 1 others. 2024.
\newblock Multimodal foundation models: From specialists to general-purpose
  assistants.
\newblock \emph{Foundations and Trends{\textregistered} in Computer Graphics
  and Vision}, 16(1-2):1--214.

\bibitem[{Li and Bansal(2023)}]{li2023panogen}
Jialu Li and Mohit Bansal. 2023.
\newblock \href {https://openreview.net/forum?id=cNObl6QQEH} {Panogen:
  Text-conditioned panoramic environment generation for vision-and-language
  navigation}.
\newblock In \emph{Thirty-seventh Conference on Neural Information Processing
  Systems (NeurIPS)}.

\bibitem[{Li et~al.(2022)Li, Tan, and Bansal}]{li2022envedit}
Jialu Li, Hao Tan, and Mohit Bansal. 2022.
\newblock Envedit: Environment editing for vision-and-language navigation.
\newblock In \emph{Proceedings of the IEEE/CVF Conference on Computer Vision
  and Pattern Recognition (CVPR)}, pages 15407--15417.

\bibitem[{Lin et~al.(2024)Lin, Nie, Wei, Chen, Ma, Han, Xu, Chang, and
  Liang}]{lin_navcot_2024}
Bingqian Lin, Yunshuang Nie, Ziming Wei, Jiaqi Chen, Shikui Ma, Jianhua Han,
  Hang Xu, Xiaojun Chang, and Xiaodan Liang. 2024.
\newblock \href {https://arxiv.org/abs/2403.07376v1} {{NavCoT}: {Boosting}
  {LLM}-{Based} {Vision}-and-{Language} {Navigation} via {Learning}
  {Disentangled} {Reasoning}}.

\bibitem[{Liu et~al.(2021)Liu, Zhu, Chang, Liang, Ge, and Shen}]{liu2021vision}
Chong Liu, Fengda Zhu, Xiaojun Chang, Xiaodan Liang, Zongyuan Ge, and Yi-Dong
  Shen. 2021.
\newblock Vision-language navigation with random environmental mixup.
\newblock In \emph{Proceedings of the IEEE/CVF International Conference on
  Computer Vision (ICCV)}, pages 1644--1654.

\bibitem[{Long et~al.(2024)Long, Li, Cai, and Dong}]{long2024discuss}
Yuxing Long, Xiaoqi Li, Wenzhe Cai, and Hao Dong. 2024.
\newblock Discuss before moving: Visual language navigation via multi-expert
  discussions.
\newblock In \emph{2024 IEEE International Conference on Robotics and
  Automation (ICRA)}, pages 17380--17387. IEEE.

\bibitem[{OpenAI(2024)}]{openai2024gpt4o}
OpenAI. 2024.
\newblock \href {https://openai.com/index/hello-gpt-4o/} {Hello gpt-4o}.

\bibitem[{Qi et~al.(2020)Qi, Wu, Anderson, Wang, Wang, Shen, and Van
  Den~Hengel}]{qi_reverie_2020}
Yuankai Qi, Qi~Wu, Peter Anderson, Xin Wang, William~Yang Wang, Chunhua Shen,
  and Anton Van Den~Hengel. 2020.
\newblock \href {https://doi.org/10.1109/CVPR42600.2020.01000} {{REVERIE}:
  {Remote} {Embodied} {Visual} {Referring} {Expression} in {Real} {Indoor}
  {Environments}}.
\newblock In \emph{2020 {IEEE}/{CVF} {Conference} on {Computer} {Vision} and
  {Pattern} {Recognition} ({CVPR})}, pages 9979--9988, Seattle, WA, USA. IEEE.

\bibitem[{Qiao et~al.(2023)Qiao, Qi, Yu, Liu, and Wu}]{qiao2023march}
Yanyuan Qiao, Yuankai Qi, Zheng Yu, Jing Liu, and Qi~Wu. 2023.
\newblock March in chat: Interactive prompting for remote embodied referring
  expression.
\newblock In \emph{Proceedings of the IEEE/CVF International Conference on
  Computer Vision}, pages 15758--15767.

\bibitem[{Radford et~al.(2021)Radford, Kim, Hallacy, Ramesh, Goh, Agarwal,
  Sastry, Askell, Mishkin, Clark et~al.}]{radford2021learning}
Alec Radford, Jong~Wook Kim, Chris Hallacy, Aditya Ramesh, Gabriel Goh,
  Sandhini Agarwal, Girish Sastry, Amanda Askell, Pamela Mishkin, Jack Clark,
  and 1 others. 2021.
\newblock Learning transferable visual models from natural language
  supervision.
\newblock In \emph{International conference on machine learning}, pages
  8748--8763. PmLR.

\bibitem[{Ramakrishnan et~al.(2021)Ramakrishnan, Gokaslan, Wijmans, Maksymets,
  Clegg, Turner, Undersander, Galuba, Westbury, Chang, Savva, Zhao, and
  Batra}]{ramakrishnan2021hm3d}
Santhosh~Kumar Ramakrishnan, Aaron Gokaslan, Erik Wijmans, Oleksandr Maksymets,
  Alexander Clegg, John~M Turner, Eric Undersander, Wojciech Galuba, Andrew
  Westbury, Angel~X Chang, Manolis Savva, Yili Zhao, and Dhruv Batra. 2021.
\newblock \href {https://arxiv.org/abs/2109.08238} {Habitat-matterport 3d
  dataset ({HM}3d): 1000 large-scale 3d environments for embodied {AI}}.
\newblock In \emph{Thirty-fifth Conference on Neural Information Processing
  Systems Datasets and Benchmarks Track}.

\bibitem[{Song et~al.(2023)Song, Wu, Washington, Sadler, Chao, and
  Su}]{song2023llmplanner}
Chan~Hee Song, Jiaman Wu, Clayton Washington, Brian~M. Sadler, Wei-Lun Chao,
  and Yu~Su. 2023.
\newblock Llm-planner: Few-shot grounded planning for embodied agents with
  large language models.
\newblock In \emph{Proceedings of the IEEE/CVF International Conference on
  Computer Vision (ICCV)}.

\bibitem[{Tan et~al.(2019)Tan, Yu, and Bansal}]{tan2019learning}
Hao Tan, Licheng Yu, and Mohit Bansal. 2019.
\newblock Learning to navigate unseen environments: Back translation with
  environmental dropout.
\newblock \emph{arXiv preprint arXiv:1904.04195}.

\bibitem[{Wang et~al.(2025{\natexlab{a}})Wang, He, Li, Qi, Hu, Yao, Liu, and
  Chen}]{wang2025clash}
Liuyi Wang, Zongtao He, Jinlong Li, Xiaoyan Qi, Mengxian Hu, Chenpeng Yao,
  Chengju Liu, and Qijun Chen. 2025{\natexlab{a}}.
\newblock \href {https://arxiv.org/abs/2512.10360} {Clash: Collaborative
  large-small hierarchical framework for continuous vision-and-language
  navigation}.
\newblock \emph{Preprint}, arXiv:2512.10360.

\bibitem[{Wang et~al.(2024)Wang, Wu, Cao, Ma, Chen, and
  Tuytelaars}]{wang-etal-2024-navigating}
Zehao Wang, Minye Wu, Yixin Cao, Yubo Ma, Meiqi Chen, and Tinne Tuytelaars.
  2024.
\newblock \href {https://doi.org/10.18653/v1/2024.findings-emnlp.269}
  {Navigating the nuances: A fine-grained evaluation of vision-language
  navigation}.
\newblock In \emph{Findings of the Association for Computational Linguistics:
  EMNLP 2024}, pages 4681--4704, Miami, Florida, USA. Association for
  Computational Linguistics.

\bibitem[{Wang et~al.(2025{\natexlab{b}})Wang, Li, Hong, Li, Li, Yu, Wang,
  Qiao, Wang, Bansal, and Wang}]{wang_bootstrapping_2024}
Zun Wang, Jialu Li, Yicong Hong, Songze Li, Kunchang Li, Shoubin Yu, Yi~Wang,
  Yu~Qiao, Yali Wang, Mohit Bansal, and Limin Wang. 2025{\natexlab{b}}.
\newblock \href {https://openreview.net/forum?id=OUuhwVsk9Z} {Bootstrapping
  language-guided navigation learning with self-refining data flywheel}.
\newblock In \emph{The Thirteenth International Conference on Learning
  Representations}.

\bibitem[{Wang et~al.(2023)Wang, Li, Hong, Wang, Wu, Bansal, Gould, Tan, and
  Qiao}]{wang2023scaling}
Zun Wang, Jialu Li, Yicong Hong, Yi~Wang, Qi~Wu, Mohit Bansal, Stephen Gould,
  Hao Tan, and Yu~Qiao. 2023.
\newblock Scaling data generation in vision-and-language navigation.
\newblock In \emph{Proceedings of the IEEE/CVF international conference on
  computer vision}, pages 12009--12020.

\bibitem[{Xiao and Zhu(2025)}]{xiao2025foundations}
Tong Xiao and Jingbo Zhu. 2025.
\newblock Foundations of large language models.
\newblock \emph{arXiv preprint arXiv:2501.09223}.

\bibitem[{Zhang et~al.(2024{\natexlab{a}})Zhang, Huang, Jin, and
  Lu}]{zhang2024vision}
Jingyi Zhang, Jiaxing Huang, Sheng Jin, and Shijian Lu. 2024{\natexlab{a}}.
\newblock Vision-language models for vision tasks: A survey.
\newblock \emph{IEEE transactions on pattern analysis and machine
  intelligence}, 46(8):5625--5644.

\bibitem[{Zhang et~al.(2025{\natexlab{a}})Zhang, Qiao, Wang, Guo, Wei, and
  Liu}]{zhang2025flexvln}
Siqi Zhang, Yanyuan Qiao, Qunbo Wang, Longteng Guo, Zhihua Wei, and Jing Liu.
  2025{\natexlab{a}}.
\newblock Flexvln: Flexible adaptation for diverse vision-and-language
  navigation tasks.
\newblock \emph{arXiv preprint arXiv:2503.13966}.

\bibitem[{Zhang et~al.(2024{\natexlab{b}})Zhang, Guo, and
  Kordjamshidi}]{zhang-etal-2024-navhint}
Yue Zhang, Quan Guo, and Parisa Kordjamshidi. 2024{\natexlab{b}}.
\newblock \href {https://aclanthology.org/2024.findings-eacl.7/} {Navhint:
  Vision and language navigation agent with a hint generator}.
\newblock In \emph{Findings of the Association for Computational Linguistics:
  EACL 2024}, pages 92--103, St. Julian's, Malta. Association for Computational
  Linguistics.

\bibitem[{Zhang and Kordjamshidi(2023)}]{zhang-kordjamshidi-2023-vln-trans}
Yue Zhang and Parisa Kordjamshidi. 2023.
\newblock \href {https://doi.org/10.18653/v1/2023.acl-long.737} {Vln-trans:
  Translator for the vision and language navigation agent}.
\newblock In \emph{Proceedings of the 61st Annual Meeting of the Association
  for Computational Linguistics (Volume 1: Long Papers)}, pages 13219--13233,
  Toronto, Canada. Association for Computational Linguistics.

\bibitem[{Zhang et~al.(2025{\natexlab{b}})Zhang, Ma, Wang, Qiao, and
  Kordjamshidi}]{zhang-2025-vln-analogical}
Yue Zhang, Tianyi Ma, Zun Wang, Yanyuan Qiao, and Parisa Kordjamshidi.
  2025{\natexlab{b}}.
\newblock \href {https://doi.org/10.18653/v1/2025.emnlp-main.759}
  {Vision-and-language navigation with analogical textual descriptions in
  {LLM}s}.
\newblock In \emph{Proceedings of the 2025 Conference on Empirical Methods in
  Natural Language Processing}, pages 15017--15025, Suzhou, China. Association
  for Computational Linguistics.

\bibitem[{Zhang et~al.(2024{\natexlab{c}})Zhang, Ma, Li, Qiao, Wang, Chai, Wu,
  Bansal, and Kordjamshidi}]{zhang2024visionandlanguage}
Yue Zhang, Ziqiao Ma, Jialu Li, Yanyuan Qiao, Zun Wang, Joyce Chai, Qi~Wu,
  Mohit Bansal, and Parisa Kordjamshidi. 2024{\natexlab{c}}.
\newblock \href {https://openreview.net/forum?id=yiqeh2ZYUh}
  {Vision-and-language navigation today and tomorrow: A survey in the era of
  foundation models}.
\newblock \emph{Transactions on Machine Learning Research}.
\newblock Survey Certification.

\bibitem[{Zhang et~al.(2025{\natexlab{c}})Zhang, Xu, Shen, Kordjamshidi, and
  Huang}]{zhang2025spartund}
Yue Zhang, Zhiyang Xu, Ying Shen, Parisa Kordjamshidi, and Lifu Huang.
  2025{\natexlab{c}}.
\newblock \href {https://openreview.net/forum?id=FGMkSL8NR0} {{SPARTUN}3d:
  Situated spatial understanding of 3d world in large language model}.
\newblock In \emph{The Thirteenth International Conference on Learning
  Representations}.

\bibitem[{Zhao et~al.(2023)Zhao, Qi, and Wu}]{zhao_mind_2023}
Chongyang Zhao, Yuankai Qi, and Qi~Wu. 2023.
\newblock \href {http://arxiv.org/abs/2308.03244} {Mind the {Gap}: {Improving}
  {Success} {Rate} of {Vision}-and-{Language} {Navigation} by {Revisiting}
  {Oracle} {Success} {Routes}}.
\newblock \emph{arXiv preprint}.
\newblock ArXiv:2308.03244 [cs].

\bibitem[{Zheng et~al.(2024)Zheng, Huang, Zhao, Zhong, and
  Wang}]{zheng2024towards_navillm}
Duo Zheng, Shijia Huang, Lin Zhao, Yiwu Zhong, and Liwei Wang. 2024.
\newblock Towards learning a generalist model for embodied navigation.
\newblock In \emph{Proceedings of the IEEE/CVF Conference on Computer Vision
  and Pattern Recognition}, pages 13624--13634.

\bibitem[{Zhou et~al.(2024{\natexlab{a}})Zhou, Li, Li, Yu, Liu, Wang, Zhang,
  Ji, Yan, He et~al.}]{zhou2024comprehensive}
Ce~Zhou, Qian Li, Chen Li, Jun Yu, Yixin Liu, Guangjing Wang, Kai Zhang, Cheng
  Ji, Qiben Yan, Lifang He, and 1 others. 2024{\natexlab{a}}.
\newblock A comprehensive survey on pretrained foundation models: A history
  from bert to chatgpt.
\newblock \emph{International Journal of Machine Learning and Cybernetics},
  pages 1--65.

\bibitem[{Zhou et~al.(2024{\natexlab{b}})Zhou, Hong, Wang, Wang, and
  Wu}]{zhou_navgpt-2_2024}
Gengze Zhou, Yicong Hong, Zun Wang, Xin~Eric Wang, and Qi~Wu.
  2024{\natexlab{b}}.
\newblock \href {https://doi.org/10.48550/arXiv.2407.12366} {{NavGPT}-2:
  {Unleashing} {Navigational} {Reasoning} {Capability} for {Large}
  {Vision}-{Language} {Models}}.
\newblock \emph{arXiv preprint}.
\newblock ArXiv:2407.12366 [cs].

\bibitem[{Zhou et~al.(2024{\natexlab{c}})Zhou, Hong, Wang, Zhao, Bansal, and
  Wu}]{zhou2024same}
Gengze Zhou, Yicong Hong, Zun Wang, Chongyang Zhao, Mohit Bansal, and Qi~Wu.
  2024{\natexlab{c}}.
\newblock Same: Learning generic language-guided visual navigation with
  state-adaptive mixture of experts.
\newblock \emph{arXiv preprint arXiv:2412.05552}.

\bibitem[{Zhou et~al.(2023)Zhou, Hong, and Wu}]{zhou_navgpt_2023}
Gengze Zhou, Yicong Hong, and Qi~Wu. 2023.
\newblock \href {https://doi.org/10.48550/arXiv.2305.16986} {{NavGPT}:
  {Explicit} {Reasoning} in {Vision}-and-{Language} {Navigation} with {Large}
  {Language} {Models}}.
\newblock \emph{arXiv preprint}.
\newblock ArXiv:2305.16986 [cs].

\end{thebibliography}

\appendix
\lstset{
  basicstyle=\ttfamily\scriptsize,  
  breaklines=true,
  frame=single,
  backgroundcolor=\color{gray!5},
  captionpos=b,
  xleftmargin=1em,
  xrightmargin=1em,
  columns=fullflexible
}

\twocolumn[%
  \begin{@twocolumnfalse}
    \vskip 0.625in minus 0.125in
    \begin{center}
        {\LARGE \bf Appendix \par}
    \end{center}
    \vskip 0.625in minus 0.125in
  \end{@twocolumnfalse}
]










\section{Skill Details}\label{sec:navnuances_appendix}

\subsection{Atomic Skill Definitions and Annotations in NavNuances}\label{sec:navnuances_atomic_skill_def}
Our atomic skills are derived from the NavNuances framework, which decomposes VLN instructions using a context-free grammar into fundamental instruction categories. In SkillNav, we adopt four atomic skills that are most directly involved in physical navigation and spatial grounding, namely Direction Adjustment, Vertical Movement, Landmark Detection, and Area and Region Identification. Each skill is constructed using strict geometric, semantic, and visibility constraints to ensure that it represents a minimal and isolatable navigation capability in the discrete VLN setting.

In SkillNav, atomic skills are defined at the semantic task level, rather than at the low-level motor execution level. This distinction is crucial for understanding what each skill represents and why they are considered atomic.

At the semantic level, an atomic skill corresponds to a single, indivisible navigation intent expressed in language, such as turning, changing floors, following a landmark, or transitioning between regions. Each skill isolates one core spatial concept and does not mix multiple independent goals.
At the execution level, the same semantic skill may be realized through multiple discrete actions in the simulator, such as several forward steps or minor orientation changes. These low-level motions are treated as motor realizations of the same semantic skill, not as separate skills.

Take ``Walk to the far end of the room'' as an example: At the semantic level, the instruction “Walk to the far end of the room” expresses a single region-based navigation intent: move to an extreme spatial location within the same room. Therefore, it is classified as one atomic skill under Area and Region Identification. At the execution level, the agent may: (1) Slightly adjust its heading. (2) Move forward across multiple viewpoints. (3) Stop near the far boundary of the room. Although multiple low-level actions are involved, these are all sub-actions serving the same semantic goal, so the instruction still activates only one atomic skill. It becomes a multi-skill instruction only when additional independent semantic goals are explicitly introduced, such as landmarks, turning commands, or vertical motion.

\paragraph{Direction Adjustment.}
This skill captures pure changes in heading direction, corresponding to actions such as turn left, turn right, and turn around.
(1) The direction change is defined by the heading angle difference between successive viewpoints.
(2) The angular change must be significant enough to indicate a clear directional shift, but not so large as to cause a U-turn or double-turn behavior, which would introduce ambiguity.
(3) We also incorporate linguistic variants of directional intent, such as “slightly turn”, “face to”, and “adjust towards”. These phrases are normalized to their effective heading change in degrees, ensuring that natural language variation does not alter the underlying atomic skill semantics.

\paragraph{Vertical Movement.}
This skill models motion across different floor levels and corresponds to go upstairs and go downstairs.
(1) Only candidates with significant elevation change (absolute height difference greater than 2 units) are considered, filtering out small slopes and near-flat transitions.
(2) The candidate viewpoint must be explicitly associated with vertical structures, such as staircases.
(3) The sign of the elevation change determines the movement type and must be consistent along the entire trajectory, for example, it is not allowed to first go upstairs and later go downstairs within a single instance.

\paragraph{Landmark Detection.}
This skill corresponds to actions such as walk towards or walk past a specific object and focuses on visually grounded object-level navigation.
(1) The viewpoint must contain salient and visually distinctive landmarks, such as sofas, desks, lamps, paintings, or large appliances, which are clearly visible.
(2) If a landmark is referenced over multiple steps, it must persist across successive views, allowing the agent to maintain a stable spatial relationship with respect to the same object. This ensures that the skill evaluates continuous landmark grounding rather than accidental encounters.

\paragraph{Area and Region Identification.}
This skill models semantic transitions between rooms or functional areas, such as go into the bedroom or exit the living room.
(1) Each trajectory must include at least one explicit region transition.
(2) Paths involving ambiguous, erroneous, or unrecognized region labels are removed or sanitized.
(3) Only horizontal region transitions are isolated, avoiding confounding factors from vertical movement. Unlike landmark-based skills, region transitions are annotated using region inclusion criteria rather than a single distance-based endpoint, reflecting the semantic nature of rooms.

We intentionally exclude Numerical Comprehension from the skill executor set because it mainly tests sequential counting and memory along corridors, rather than continuous motion control. It is instead handled implicitly by the higher-level routing module when needed.

These four skills are considered ``atomic'' because they correspond to the lowest-level, non-decomposable navigation operations in the discrete VLN setting. 
They each focus on one isolated capability.
They do not mix multiple semantic or geometric concepts. Each instruction in NavNuances is generated to isolate exactly one such capability using the CFG production rules.

In the R2R-style discrete viewpoint selection setting, these skills are operationally independent. For example, for the sub-goal “Go through the double doors”, only the Landmark Recognition skill is activated. The agent directly selects the candidate viewpoint that contains the double doors in its panoramic observation, and then traverses them. This execution does not rely on Direction Adjustment, because the simulator action is viewpoint selection rather than continuous steering.

\subsection{Explanation on Lowest Success Rate of Landmark Recognition in NavNuances}\label{sec:navnuances_landmark_failure}
The relatively low performance in the Landmark Recognition (LR) category mainly stems from the fine-grained nature of both the task and its evaluation. First, LR requires instance-level grounding rather than category-level recognition. The evaluation protocol does not merely check whether an object category is detected, but whether the agent’s final position satisfies precise geometric constraints with respect to the specific landmark instance, such as reducing the distance to the object center for “walk towards,” or truly crossing the object for “walk past.” This level of precision goes well beyond the coarse object semantics typically learned in VLN training.

Second, annotation difficulty and viewpoint sparsity further increase the challenge. Due to the discrete navigation graph and limited viewpoints in Matterport3D environments, landmarks are often visible only from a narrow range of poses. Even small deviations in viewpoint selection can cause large geometric errors under the distance-based evaluation, which amplifies minor perception or control mistakes into outright failures.

Third, models frequently struggle with the spatial relational concept of “walk past.” As shown by failure cases, agents often stop beside the object and mistakenly interpret proximity as having passed it, or confuse front and back views of the landmark. This indicates that the bottleneck is not simply recognizing the object but understanding the temporal and geometric meaning of passing an object, which requires consistent spatial reasoning across multiple steps.

Finally, we hypothesize that the main limitation is not low-level object detection but higher-level spatial reasoning and decision-making. This is supported by the observation that stronger vision-language models substantially improve the “walk towards” subset, where recognition dominates, but still struggle with “walk past,” where reasoning about relative motion and spatial relations is critical. This directly motivates our design choice of using a VLM as a routing and reasoning module, where rich visual perception is explicitly coupled with instruction-conditioned geometric and commonsense reasoning, rather than relying solely on perception-driven navigation policies.

\section{Primary Factors of Trajectory Generation}\label{sec:data_synthesis_appendix}



\begin{figure}
    \centering
    \includegraphics[width=\linewidth]{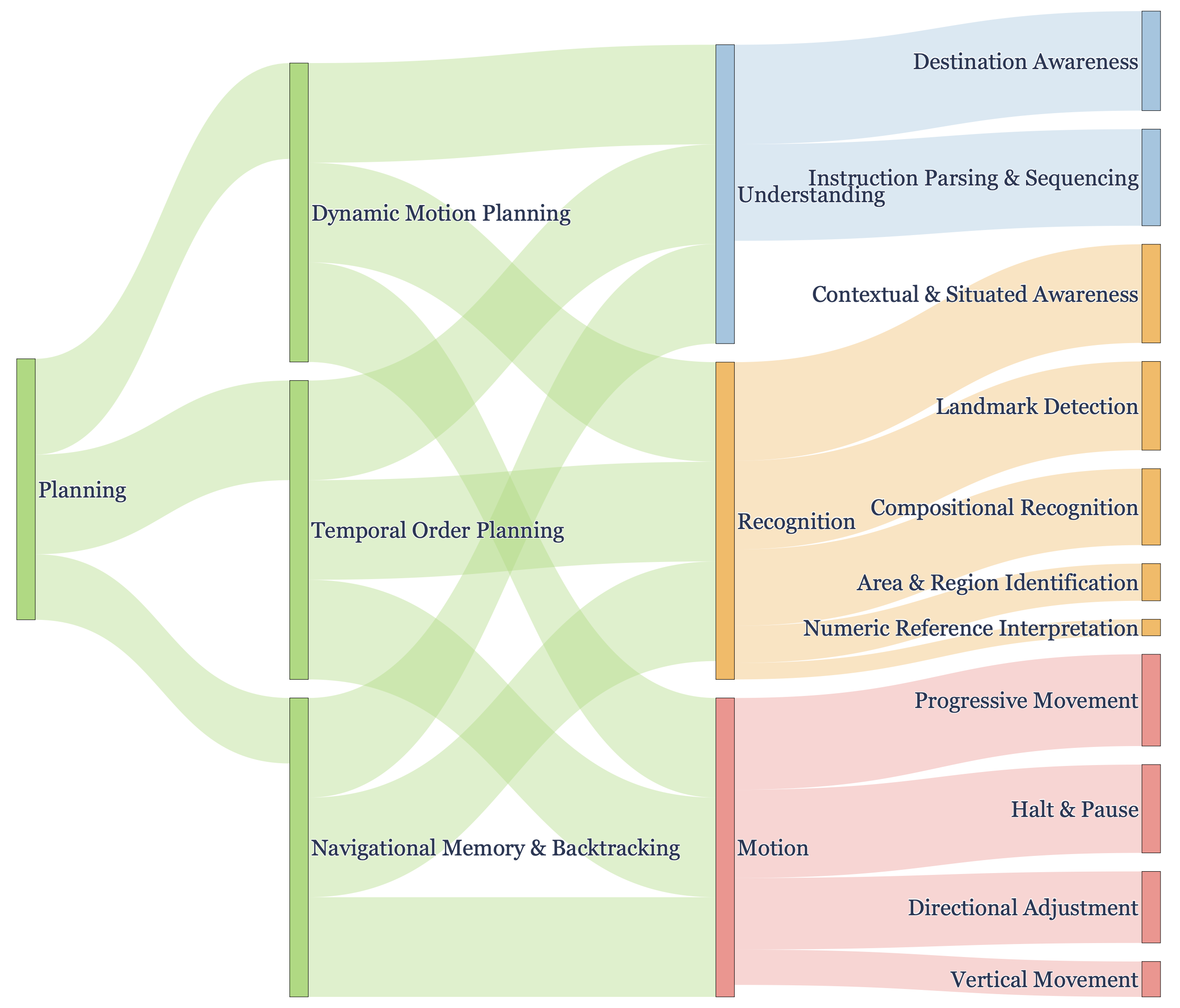}
    \caption{Skill Collection for the Skill Taxonomy. we categorize navigation into \textit{High-level Planning} (System 2, e.g., Temporal Order, Navigational Memory) and \textit{Low-level Execution} (System 1, e.g., Motion primitives, Landmark Recognition). }
    \label{fig:skill_collection}
\end{figure}

As introduced in Section Skill-Specific Data Synthesis and Agent Training in Methodology, we construct 6 skill-specific datasets and train the agents based on them. The primary factors for the construction of each skill are as follows:



\paragraph{Temporal Order Planning.} (1) A random initial move is selected. (2) Staying in the same region (\eg, hallway $\rightarrow$ hallway) for the first half of the trajectory to encourage temporal continuity at first. (3) Once halfway through, the agent is allowed (and encouraged) to transition to new regions.

\paragraph{Direction Adjustment.} (1) The direction change is based on the heading degree. (2) It should be significant enough to indicate a directional shift, but not so large as to cause a reversal or double-turn behavior. 

\paragraph{Vertical Movement.} (1) Only candidates with significant elevation (more than $\pm 2$) are considered, which filters out nearly flat or slight inclines/declines. (2) The candidate viewpoint must be explicitly marked as vertically relevant (\eg, stairs). (3) The elevation sign determines movement type, and it must be consistent with the applied trajectory. For instance, it is impossible to go upstairs and then go downstairs in one case.

\begin{figure}[!ht]
    \centering
    \includegraphics[width=\linewidth]{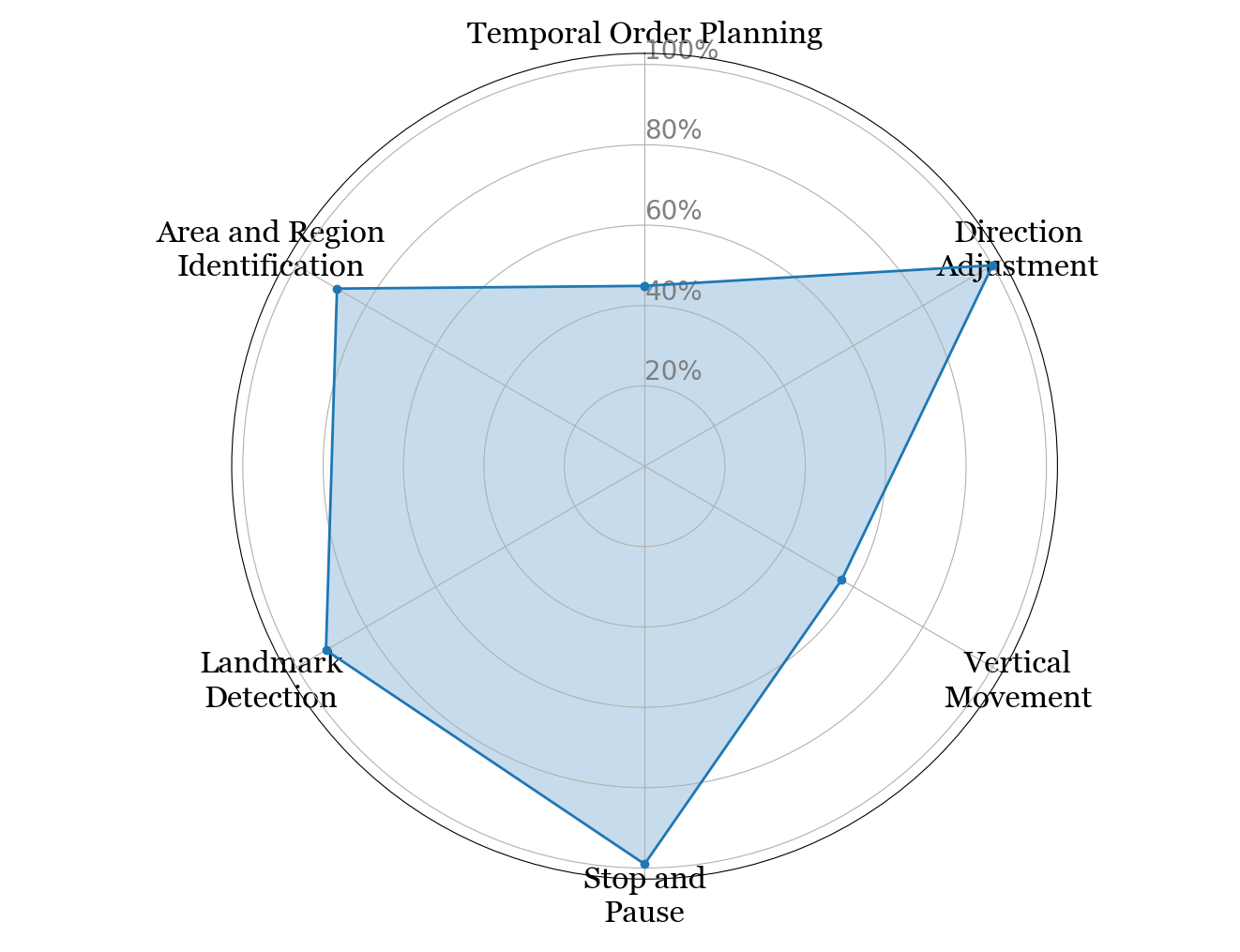}
        \caption{Distribution of instructions in the R2R dataset categorized by the proposed skill taxonomy.
        }
    \label{fig:skill_instruction_distribution}
\end{figure}

\paragraph{Stop and Pause.} (1) The stop should occur at a place with or after semantically relevant context for pausing, \eg, in front of a painting, at the foot of stairs. (2) The candidate image is very similar to the previous viewpoints. 

\paragraph{Landmark Detection.} (1) The viewpoint must include obvious, visually distinctive landmarks or objects (\eg, sofa, desk, painting, lamp) clearly visible in the image. (2) If a landmark is to be referenced over multiple steps, it should appear persistently in successive views, allowing the agent to maintain spatial awareness relative to that object.

\paragraph{Area and Region Identification.} (1) A trajectory must include at least one region change. (2) Paths with ``Error'' or unrecognized regions are ignored or sanitized. (3) All horizontal region changes are isolated.

%
%

\section{Skill Taxonomy and Sensitivity}\label{sec:appendix_skill_taxonomy_sensitivity}

The current SkillNav executor library is designed as a compact set of semantic navigation primitives rather than a closed action inventory. Each skill captures one primary instruction intent at the semantic level, while the simulator may realize that intent through several low-level viewpoint transitions. For example, ``walk to the far end of the room'' is treated as an Area and Region Identification instruction even though execution may involve multiple forward moves and minor heading corrections. The instruction becomes multi-skill only when it introduces another independent semantic goal, such as an explicit turn, a landmark trigger, or a vertical transition.

To test whether the selected skill set is useful as a set rather than merely a larger ensemble, we evaluate subsets of the five skill executors on R2R Val-Unseen. The full five-executor setting performs best, while several four-executor and three-executor variants are weaker. This indicates that the gain comes from precise delegation to specialized heads rather than simply increasing the number of available agents.

\begin{table}[h]
\centering
\caption{Sensitivity analysis on skill set size using R2R Val-Unseen. The weight vector follows $(\pi_{\text{da}}, \pi_{\text{vm}}, \pi_{\text{sp}}, \pi_{\text{ld}}, \pi_{\text{ar}})$ for Direction Adjustment, Vertical Movement, Stop and Pause, Landmark Detection, and Area/Region Identification.}
\label{tab:skill_set_sensitivity}
\resizebox{\linewidth}{!}{
\begin{tabular}{cccc}
\toprule
\textbf{\# Agents} & \textbf{Weight} & \textbf{SR (\%) $\uparrow$} & \textbf{SPL (\%) $\uparrow$} \\
\midrule
2 & $(1,0,0,0,1)$ & $79.57$ & $70.30$ \\
2 & $(1,1,0,0,0)$ & $78.71$ & $69.54$ \\
3 & $(0,1,0,1,1)$ & $79.78$ & $70.43$ \\
3 & $(1,1,1,0,0)$ & $79.27$ & $69.89$ \\
4 & $(0,1,1,1,1)$ & $80.80$ & $71.29$ \\
4 & $(1,1,1,0,1)$ & $79.23$ & $70.13$ \\
5 & $(1,1,1,1,1)$ & $\mathbf{82.59}$ & $\mathbf{76.50}$ \\
\bottomrule
\end{tabular}
}
\end{table}

This result does not imply that the current taxonomy is universally exhaustive. Instead, it supports the current five-executor library as an effective basis for the evaluated indoor VLN setting. For environments with new capabilities, such as object manipulation or human-aware navigation, SkillNav can be extended by training an additional executor and adding its natural-language skill definition to the router prompt.

\section{Path Length in Trajectory Generation}\label{sec:trajectory_generation_appendix}
\begin{figure*}[!ht]
    \centering
    \includegraphics[width=0.9\linewidth]{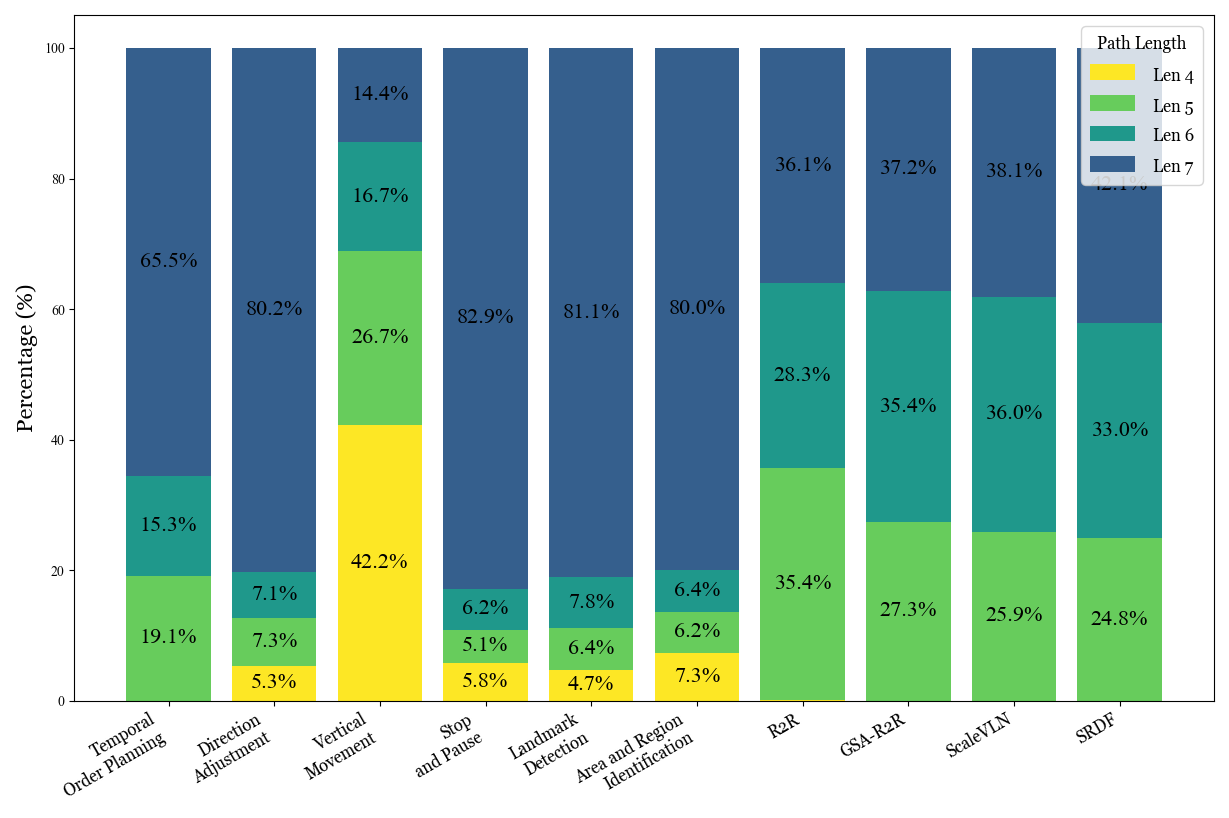}
    \caption{The statistics of the path length of our synthetic datasets compared with existing VLN datasets. The R2R, ScaleVLN, SRDF datasets, and our 6 skill-specific datasets are all for training, while only GSA-R2R is for evaluation.}
    \label{fig:path_length_distribution}
\end{figure*}

We constrain trajectory length to 4–7 steps to keep the difficulty and temporal context comparable to natural VLN data. 
Figure~\ref{fig:path_length_distribution} shows the statistics of the path length.
To be noted, the R2R, ScaleVLN, SRDF datasets, and our Temporal Order Planning datasets have significantly fewer instructions with a 4-step trajectory.



\section{Synthetic Data Integrity and Leakage Analysis}\label{sec:synthetic_data_bias_learning}
We analyze our synthetic data to test whether agents learn meaningful semantics rather than overfitting to linguistic triggers, and to verify that the datasets do not leak information into evaluation splits. 

\subsection{Skill Keyword Coverage}
We randomly sampled 100 instructions from each skill-specific dataset and manually extracted high-frequency keywords or phrases associated with motion, regions, and objects.
The utilization of GPT-4o for instruction generation during dataset construction introduces linguistic diversity, expanding the vocabulary beyond the R2R and GSA-R2R datasets. For instance, the Direction Adjustment synthetic dataset includes complex phrases such as ``adjust your direction to align with...''.

Human evaluation on the synthetic datasets reveals that keyword presence is not a reliable proxy for action, preventing agents from overfitting to specific words. A notable example is the preposition ``down'. In the instruction ``proceed down the hallway', the term ``down' implies forward progression from one end to the other, whereas in vertical contexts, it denotes an elevation change. This semantic ambiguity forces the agent to rely on visual context and sentence structure rather than mapping the word ``down'' exclusively to a vertical motion primitive.

Table~\ref{tab:keyword_coverage} reports keyword coverage across datasets. Each synthetic dataset exhibits a dominant proportion of keywords associated with its target skill, confirming the effectiveness of targeted data generation. At the same time, substantial linguistic overlap and semantic complexity across datasets encourage the learning of robust, context-aware policies instead of dataset-specific shortcuts.

The Vertical Movement dataset provides a particularly informative case. As shown in Table~\ref{tab:keyword_coverage}, it contains a high density of non-vertical concepts, including Landmark Object keywords ($18.72\%$) and Direction Motion keywords ($8.05\%$), compared with only $3.32\%$ Vertical Motion keywords. This reflects the physical nature of vertical navigation, which necessarily involves horizontal motion relative to landmark structures. Despite this lexical overlap, the Vertical Movement dataset remains the primary source of Vertical Motion keywords, with less than $0.2\%$ in all other datasets. Moreover, all trajectories in this dataset are strictly filtered to ensure consistent elevation changes in the simulator. As a result, the agent cannot rely on generic movement words or visual objects such as stairs as direct triggers for vertical behavior. Instead, performance gains arise from the precise alignment between linguistic cues, visual affordances, and trajectory constraints. This alignment encourages the learning of a robust compositional policy rather than a brittle keyword-to-action mapping.

\begin{table*}[t]
\centering
\caption{Keyword coverage analysis across synthetic datasets. The coverage is measured by the proportion of words in the dataset that match the target skill keywords. Bold indicates the highest coverage for each keyword category. Ident.= Identification.}
\label{tab:keyword_coverage}
\begin{tabular}{lccccc}
\toprule
\textbf{Dataset} & \textbf{Direction} & \textbf{Vertical} & \textbf{Stop} & \textbf{Landmark} & \textbf{Region} \\
\textbf{ / Keyword} & \textbf{Motion} & \textbf{Motion} & \textbf{Motion} & \textbf{Object} & \textbf{Type} \\
\midrule
Direction Adjustment   & \textbf{14.56\%} & 0.03\%          & 0.06\%           & 13.64\%          & 9.62\%           \\

Vertical Movement      & 8.05\%           & \textbf{3.32\%} & 0.01\%           & 18.72\%          & 9.46\%           \\

Stop and Pause         & 12.56\%          & 0.18\%          & \textbf{18.12\%} & 18.02\%          & 9.57\%           \\

Landmark Detection     & 9.53\%           & 0.11\%          & 0.97\%           & \textbf{27.88\%} & 7.58\%           \\

Area and Region Ident. & 7.86\%           & 0.05\%          & 0.10\%           & 16.39\%          & \textbf{19.89\%} \\
\bottomrule
\end{tabular}
\end{table*}

\subsection{Skill Agents to Skill Synthetic Datasets}

To further assess specialization and generalization, we evaluate each skill-specific agent on both its own synthetic dataset and the other skill-specific datasets. Table~\ref{tab:skill_2_skills} reports the success rate (SR) and success weighted by path length (SPL) for all cross-skill evaluations. Each agent achieves its strongest performance on the dataset corresponding to its trained skill, while maintaining non-trivial performance on the remaining datasets. This pattern indicates that although the agents develop skill-specific expertise, they do not collapse into narrowly overfit behaviors.

\begin{table*}[h!]
\centering
\caption{Each synthetic subset reports both SR and SPL for agents trained and tested on synthetic instructions. Bold indicates the highest result in each column.}
\label{tab:skill_2_skills}
\begin{tabular}{l*{5}{cc}}
\toprule
\multirow{2}{*}{\textbf{Agent}} &
\multicolumn{2}{c}{\begin{tabular}{c}\textbf{Direction} \\ \textbf{Synthetic}\end{tabular}} &
\multicolumn{2}{c}{\begin{tabular}{c}\textbf{Vertical} \\ \textbf{Synthetic}\end{tabular}} &
\multicolumn{2}{c}{\begin{tabular}{c}\textbf{Stop} \\ \textbf{Synthetic}\end{tabular}} &
\multicolumn{2}{c}{\begin{tabular}{c}\textbf{Landmark} \\ \textbf{Synthetic}\end{tabular}} &
\multicolumn{2}{c}{\begin{tabular}{c}\textbf{Region} \\ \textbf{Synthetic}\end{tabular}} \\
\cmidrule(lr){2-3} \cmidrule(lr){4-5} \cmidrule(lr){6-7} \cmidrule(lr){8-9} \cmidrule(lr){10-11}
& \textbf{SR} & \textbf{SPL}
& \textbf{SR} & \textbf{SPL}
& \textbf{SR} & \textbf{SPL}
& \textbf{SR} & \textbf{SPL}
& \textbf{SR} & \textbf{SPL} \\
\midrule
Direction DUET & \textbf{49.33} & \textbf{43.35} & 38.89 & 30.77 & 41.11 & 36.44 & 38.44 & 34.12 & 35.56 & 30.92 \\
Vertical DUET  & 35.56 & 31.11 & \textbf{65.11} & \textbf{55.65} & 39.78 & 35.96 & 37.78 & 33.69 & 36.00 & 30.72 \\
Stop DUET      & 36.22 & 31.79 & 34.67 & 27.37 & \textbf{52.67} & \textbf{48.55} & 42.22 & 38.76 & \textbf{42.22} & 36.12 \\
Landmark DUET  & 36.00 & 30.46 & 46.00 & 38.35 & 41.11 & 36.69 & \textbf{44.22} & \textbf{39.98} & 38.67 & 34.02 \\
Region DUET    & 34.00 & 28.89 & 49.56 & 41.51 & 38.44 & 33.97 & 36.44 & 32.90 & 36.67 & \textbf{43.02} \\
\bottomrule
\end{tabular}
\end{table*}

\subsection{Skill to Temporal Subsets}
We additionally conduct cross-style evaluation on each skill-based agent using synthetic temporal datasets. For this setting, each agent is trained on its skill-specific synthetic data combined with the temporal order planning synthetic data, and then evaluated on four temporal subsets filtered from R2R Val Unseen. These subsets capture different forms of temporal reasoning, including Conditional Immediacy, Duration Bound, Forward Sequential, and Reverse Sequential instructions.



Table~\ref{tab:skill_2_temporal_subset} presents SR and SPL on the four temporal subsets. All evaluations are performed on synthetic instructions that are specifically designed to remove superficial linguistic cues. The consistent performance across temporal subsets demonstrates that agents trained with skill-specific synthetic data retain strong compositional and temporal reasoning abilities, rather than overfitting to surface-level instruction patterns.

\begin{table*}[h!]
\centering
\caption{Each temporal subset reports both SR and SPL, and all agents are trained and tested on synthetic instructions designed to remove superficial linguistic cues. Bold indicates the highest result for each metric within the subset.}
\label{tab:skill_2_temporal_subset}
\begin{tabular}{l*{4}{cc}}
\toprule
\multirow{2}{*}{\textbf{Agent}} &
\multicolumn{2}{c}{\begin{tabular}{c}\textbf{Conditional} \\ \textbf{Immediacy}\end{tabular}} &
\multicolumn{2}{c}{\begin{tabular}{c}\textbf{Duration} \\ \textbf{Bound}\end{tabular}} &
\multicolumn{2}{c}{\begin{tabular}{c}\textbf{Forward} \\ \textbf{Sequential}\end{tabular}} &
\multicolumn{2}{c}{\begin{tabular}{c}\textbf{Reverse} \\ \textbf{Sequential}\end{tabular}} \\
\cmidrule(lr){2-3} \cmidrule(lr){4-5} \cmidrule(lr){6-7} \cmidrule(lr){8-9}
& \textbf{SR} & \textbf{SPL} & \textbf{SR} & \textbf{SPL} & \textbf{SR} & \textbf{SPL} & \textbf{SR} & \textbf{SPL} \\
\midrule                                                                            
ScaleVLN       & 84.29 & 76.29 & 76.27 & 67.45 & 79.53 & 68.92 & 74.29 & 66.97 \\
Temporal DUET  & \textbf{88.57} & \textbf{82.18} & \textbf{84.18} & 74.90 & 85.83 & 76.93 & \textbf{88.57} & \textbf{81.72} \\
Direction DUET & 84.29 & 77.57 & 83.05 & \textbf{76.69} & 85.83 & 77.99 & 84.29 & 76.91 \\
Vertical DUET  & 86.43 & 79.71 & 79.66 & 74.31 & 86.61 & 80.35 & 80.00 & 75.08 \\
Stop DUET      & 85.00 & 79.91 & 81.92 & 74.59 & 87.40 & 80.34 & 82.86 & 77.15 \\
Landmark DUET  & 87.14 & 81.58 & 80.79 & 75.43 & \textbf{88.19} & \textbf{81.47} & 81.43 & 77.72 \\
Region DUET    & 85.00 & 80.06 & 79.66 & 72.93 & 87.40 & 80.42 & 74.29 & 69.86 \\
\bottomrule
\end{tabular}
\end{table*}

\subsection{Data Leakage Assessment}
We evaluate whether GSA-R2R gains could be driven by data leakage from shared scenes or instruction augmentation. Two potential channels are considered: (1) \emph{vision-level scene overlap} through HM3D-based augmentation and (2) \emph{language-level instruction overlap} between training data and the GSA-R2R test sets. HM3D overlap is shared by all competing methods and does not uniquely benefit SkillNav, and we find no meaningful language-level leakage from either ScaleVLN or our synthetic datasets.


\paragraph{Vision-Level Scene Data Overlap with HM3D.}




Each SkillNav agent is fine-tuned with ScaleVLN augmentation data that includes 899 HM3D scans; GSA-R2R also contains HM3D scenes. SRDF and ScaleVLN use the same HM3D-based augmentation, so any scene-level overlap affects all methods equally rather than uniquely benefiting SkillNav. Our synthetic skill trajectories are constructed only from the seen split of MP3D R2R, not from GSA-R2R, preventing direct trajectory or instruction leakage.

\paragraph{No Language-Level or Instruction Leakage.}


We first analyze the vocabulary overlap between the \textbf{ScaleVLN augmentation instructions} and the \textbf{GSA-R2R test instructions}, since ScaleVLN is used for vision-level fine-tuning and is the primary potential source of language leakage. 

All instructions are lowercased and tokenized using the regular expression (\texttt{[a-zA-Z']+}) to construct unique vocabularies. Under this protocol, the ScaleVLN augmentation file contains \textbf{208 unique tokens}. Although ScaleVLN includes 2,891,134 instruction entries, these instructions are generated by a constrained template-based pipeline with a fixed pool of action verbs, spatial prepositions, and object categories. As a result, the number of unique token types remains small despite the large number of instruction instances.

\textbf{ScaleVLN vs. GSA-R2R Vocabulary Overlap.}
Table~\ref{tab:scalevln_gsa_vocab} reports the vocabulary overlap between ScaleVLN augmentation instructions and the three GSA-R2R test splits. While ScaleVLN covers a large fraction of its own limited vocabulary in GSA-R2R, the coverage of the GSA-R2R test vocabulary by ScaleVLN remains low. In particular, across the union of all three GSA-R2R test sets, only 196 out of 1,645 tokens overlap, corresponding to \textbf{11.9\% test vocabulary coverage}. This indicates that the majority of GSA-R2R test words are not observed in ScaleVLN augmentation.

\begin{table}[h]
\centering
\caption{Vocabulary overlap between ScaleVLN augmentation instructions and GSA-R2R test sets.}
\label{tab:scalevln_gsa_vocab}
\resizebox{\linewidth}{!}{
\begin{tabular}{lccc}
\toprule
\textbf{GSA-R2R Test Split} & \textbf{Overlap} & \textbf{Test Vocab Size} & \textbf{Coverage (\%)} \\
\midrule
Residential Basic & 186 & 514 & 36.2 \\
Non-Residential Basic & 166 & 578 & 28.7 \\
Non-Residential Scene & 168 & 1,376 & 12.2 \\
\midrule
Union of All Tests & 196 & 1,645 & 11.9 \\
\bottomrule
\end{tabular}
}
\end{table}


We also report the vocabulary coverage of our GPT-4o-generated synthetic skill datasets with respect to the union of all GSA-R2R test vocabularies. The atomic skill coverage ranges from \textbf{21.95\% to 27.72\%} across different skill types.
Even in the highest case (Temporal Order Planning skill), more than 58\% of the GSA-R2R test vocabulary remains unseen. 
This confirms that our synthetic language does not duplicate the GSA-R2R test distribution.


Taken together, these results demonstrate that the performance gains on GSA-R2R cannot be explained by either vision-level or language-level data leakage from HM3D, ScaleVLN, or our synthetic annotations. Instead, the improvements stem from the proposed skill decomposition and compositional learning framework.

\section{Reproducibility Details}\label{sec:appendix_reproducibility}

We evaluate two model configurations: the standard ScaleVLN setup and the SRDF-augmented SkillNav variant. Both architectures utilize ViT-B/16 to extract $768$-dimensional object features and adhere to an identical two-stage training protocol.

For the ScaleVLN configuration, we employ CLIP-B/16~\citep{radford2021learning} as the visual encoder with a feature dimension of 512 and BERT-base-uncased~\citep{bert-base-uncased} as the language backbone, while utilizing ViT-B/16 to extract object features with a dimension of 768. 
During the skill training, we fine-tune the DUET pre-trained model with Temporal Order synthetic data, ScaleVLN augmentation data, and R2R Train data for $50,000$ iterations using a batch size of $32$ and a learning rate of \( 5 \times 10^{-5}\) on 1 NVIDIA A6000 GPU with the random seed $0$. The best finetuned Temporal DUET model is selected based on the SPL performance on the R2R Validation Unseen dataset. Based on the Temporal DUET, we employ the second round fine-tuning with atomic skill synthetic data for $30,000$ iterations with a batch size of $16$ on the same GPU, a protocol that the SRDF-augmented SkillNav variant also follows regarding optimizer, learning rate, and iteration budget.

For the SRDF-augmented SkillNav variant, we replace the visual encoder with InternViT-6B and set the feature dimension as 3200 while retaining the same 768-dimension object features from ViT-B/16. Training follows the same two-stage recipe as the ScaleVLN variant: (1) 50,000 iterations on Temporal Order synthetic data, ScaleVLN augmentation, and R2R Train with AdamW, batch size $32$, and learning rate \(5 \times 10^{-5}\); (2) 30,000 iterations on atomic skill synthetic data with batch size $16$. Both stages run on a single NVIDIA A6000 GPU with random seed $0$, keeping optimizer and hyperparameters identical across variants.

The router is run with a maximum context length of 40,960 tokens, a temperature of 0, and greedy decoding. All router inferences are executed under the vLLM framework using five beam search candidates. The routing decision is performed in \texttt{top1} mode with \texttt{argmax} feedback. Instruction reordering is enabled, and routing weights are treated as integers and normalized before use.

All experiments are evaluated in test-only mode with detailed output and submission enabled. Multiple expert checkpoints are loaded through \texttt{resume\_files}, with their corresponding normalized \texttt{resume\_weights}. The external VLM router is accessed via a dedicated server URL specified by \texttt{router\_server\_url}. The feature backbone uses CLIP ViT-B/16 with 512-dimensional visual features, and all evaluations are conducted with a fixed random seed $0$ for deterministic behavior. Batch size is explicitly specified in the execution scripts for each model configuration.

To support reproducibility of the routing and fallback behavior, we release the exact action routing prompt template used for Qwen2.5-VL-7B-Instruct and GLM-4.1v-9B-Thinking in Appendix~\ref{sec:appendix_prompt_action_router}, Listing~\ref{lst:subgoal_localizer} and Listing~\ref{lst:skill_router}. 


\paragraph{Artifact Use.}
We use R2R, GSA-R2R, NavNuances, Matterport3D/HM3D, DUET, ScaleVLN, SRDF, GPT-4o, Qwen2.5-VL, GLM, and InternVL as research artifacts for benchmark evaluation, feature extraction, navigation-policy initialization, synthetic instruction generation, and routing. Our use follows their intended academic research roles: public VLN benchmarks are used for training/evaluation, prior VLN models are used as baselines or initialization backbones, and LLM/VLM systems are used for data synthesis and inference-time routing. Users of released code or data should follow the corresponding dataset, model, and API licenses or terms of use.

\section{Temporal Order Planning Agent}\label{sec:temporal_order_planning_agent}

As introduced earlier, the training of each skill-based agent follows a two-stage fine-tuning strategy. 
In the first stage, we fine-tune a pre-trained DUET model using a combination of the R2R training split, ScaleVLN augmentation data, and our proposed Temporal Synthetic dataset, resulting in a strong skill-agnostic backbone. 
We evaluate this first-stage model on the R2R Val Unseen split across four temporal logic subsets.

Temporal Order Planning captures the agent’s ability to reason over the sequence and structure of subgoals. 
Compared to ScaleVLN, our model demonstrates improved temporal reasoning capabilities, as detailed in Table~\ref{tab:temporal_subset_performance}. 
This improvement comes from enhanced \textbf{Temporal Order Planning}, which enables the agent to reason about the sequence and structure of subgoals. The Temporal Order Planning subsets include:
\begin{itemize}
    \item \textbf{Conditional immediacy}: The agent must execute an action immediately after a specific condition is met. These instructions are typically triggered by phrases such as \textit{once}, \textit{as soon as}, or \textit{upon}. (\eg, ``Once you enter the hallway, turn left'')
    
    \item \textbf{Bounded duration}: The agent is required to maintain an action until a specific condition becomes true. These instructions use keywords such as \textit{until} or \textit{while}. (\eg, ``Keep walking until you see the staircase'')
    
    \item \textbf{Forward sequential}: These instructions describe a sequence where Action B follows Action A in order. Temporal cues include \textit{then}, \textit{finally}, \textit{before}, and \textit{after}. (\eg, ``Go forward, then turn right, and finally stop'')
    
    \item \textbf{Backward sequential}: Action B is described first but should occur only after Action A. These often use similar cues as (\eg, ``Before turning, make sure you’re at the hallway entrance''), but the order of mention and execution differs.
\end{itemize}




Unlike low-level action chaining, temporal order planning involves higher-level temporal logic that determines when and how atomic skills should be executed in sequence.
As shown in Table~\ref{tab:temporal_subset_performance}, our Temporal Synthetic Data improves navigation in failure cases where prior methods such as ScaleVLN struggle.

\begin{table*}[t]
\centering
\caption{Navigation performance across 4 temporal logic instructions from R2R Val Unseen dataset.
\textbf{Bold} values denote metrics that exceed the R2R Val Unseen average, while 
\textcolor{gray}{gray} values indicate metrics that fall below the average.
Temporal DUET is the agent fine-tuned with the Temporal Order Planning synthetic dataset in the first training stage.
}
\label{tab:temporal_subset_performance}

\resizebox{\linewidth}{!}{

\begin{tabular}{lcccccc}
\toprule
\textbf{Model} &
\begin{tabular}{c} \textbf{Val} \\ \textbf{Unseen} \end{tabular} &
\begin{tabular}{c} \textbf{Conditional} \\ \textbf{Immediacy} \end{tabular} &
\begin{tabular}{c} \textbf{Bounded} \\ \textbf{Duration} \end{tabular} &
\begin{tabular}{c} \textbf{Forward} \\ \textbf{Sequential} \end{tabular} &
\begin{tabular}{c} \textbf{Backward} \\ \textbf{Sequential} \end{tabular} \\
\midrule

\textbf{ScaleVLN (SR)} 
  & 79.23 
  & \textbf{84.29} 
  & \textcolor{gray}{76.27} 
  & \textbf{79.53} 
  & \textcolor{gray}{74.29} \\

\textbf{Temporal DUET (SR)} 
  & 82.59 
  & \textbf{88.57} 
  & \textbf{84.18} 
  & \textbf{85.83} 
  & \textbf{88.57} \\
\midrule

\textbf{ScaleVLN (SPL)} 
  & 69.96 
  & \textbf{76.29} 
  & \textcolor{gray}{67.45} 
  & \textcolor{gray}{68.92} 
  & \textcolor{gray}{66.97} \\

\textbf{Temporal DUET (SPL)} 
  & 75.40 
  & \textbf{82.18} 
  & \textcolor{gray}{74.90} 
  & \textbf{76.93} 
  & \textbf{81.72} \\
\bottomrule
\end{tabular}
}
\end{table*}

\section{Stop and Pause Agent}\label{sec:stop_agent}

The Stop and Pause agent integrates two stopping mechanisms within the DUET framework: (1) the agent can explicitly issue a stop action at a given viewpoint; and (2) if the agent does not explicitly stop when the navigation loop ends, DUET retrospectively selects the visited location with the highest stop probability and optionally appends a shortest path to reach it. Since we apply a stopping-focused data augmentation strategy that exposes the model to diverse stop-relevant cues during training, this supervision enables the agent to distinguish between the two stopping mechanisms and to learn when stopping aligns with the instruction intent and visual context. Although NavNuances does not include a dedicated stopping split, our Stop agent still outperforms generalist baselines like ScaleVLN and SRDF across all skill categories (Table~\ref{tab:performance_navnuance}), suggesting that effective stopping is a foundational capability that influences the success of diverse navigation behaviors.

\section{Skill Routing Effectiveness}\label{sec:appendix_skill_routing}

\subsection{Action Router Ablations}\label{sec:appendix_action_router_ablation}

As shown in Table~\ref{tab:temporal_reorder_ablation}, the Random baseline achieves a high success rate of $78.39\%$ on Test-R-Basic, confirming that the underlying skill-based agents are inherently capable of executing navigation subgoals. However, the necessity of active reasoning becomes evident in the Test-N-Scene split. In these novel environments, VLM-based routers consistently outperform the random baseline, proving that semantic understanding is crucial for correctly routing instructions to skills when the scene is unknown. Among the models, GPT-4o demonstrates the strongest generalization, achieving the highest Test-N-Scene performance with $\mathbf{58.16}\%$ SR and $\mathbf{48.96}$ SPL. This indicates that while the base agents provide a solid foundation, advanced reasoning models are the key differentiator for handling the complexity of unseen scenes.




\subsection{Skill Transition}

Regarding inter-skill interaction and interference, we computed the transition probability matrix $P(\text{next} \mid \text{current})$ to map temporal dependencies between skill. The data suggests cooperative chaining rather than negative interference. For instance, Landmark Detection has a very high probability of transitioning to Stop and Pause ($0.3516$), indicating that the agent correctly utilizes landmarks as verification signals to identify the goal location. Similarly, Direction Adjustment frequently leads to Area and Region Identification ($0.3916$), suggesting a behavior where the agent re-orients itself physically before re-evaluating the semantic context of the room. Notably, the transition from Landmark to Direction is $0.00$, implying that once a specific landmark is identified, the agent prioritizes verification (stopping) or vertical navigation over immediate rotational correction.

We analyze the skill routing in R2R Val Unseen of SkillNav with Qwen2.5-VL-7B in the following table
\begin{table*}[h!]
\centering
\caption{
Inter skill transition probabilities $P(\text{next} \mid \text{current})$.  
}
\begin{tabular}{lccccc}
\hline
 & \textbf{Direction} & \textbf{Vertical} & \textbf{Stop} & \textbf{Landmark} & \textbf{Region} \\
\hline
\textbf{Direction} 
  & 0.3449 & 0.1971 & 0.1374 & 0.2108 & 0.3916 \\
\textbf{Vertical}  
  & 0.3415 & 0.2414 & 0.1729 & 0.0931 & 0.1035 \\
\textbf{Stop}      
  & 0.0610 & 0.2385 & 0.2296 & 0.1854 & 0.2984 \\
\textbf{Landmark}  
  & 0.0000 & 0.2044 & 0.3516 & 0.2711 & 0.0280 \\
\textbf{Region}    
  & 0.2525 & 0.1186 & 0.1084 & 0.2397 & 0.1785 \\
\hline
\end{tabular}

\end{table*}





\subsection{Human Evaluation of Skill Routing}\label{sec:appendix_human_eval_skill_routing}

To address the concern regarding the routing mechanism's decision-making process, we conducted a detailed human evaluation of failure cases in the R2R Val-Unseen split. By analyzing the specific steps where the agent failed, we provide evidence that the router is not behaving as a ``black box'', but rather as a linguistically competent planner that is primarily bottlenecked by visual grounding limitations.

We examined 17 distinct failure cases where specific step-level data was available. Quantitatively, the average success rate of the routing mechanism, defined as the proportion of steps completed successfully before the first failure, is 24.5\%:
\[ \frac{1}{N} \sum \frac{\text{first\_failure\_step} - 1}{\text{\#routing\_steps}} \]
This indicates that while the agent effectively initiates navigation, critical routing errors tend to occur in the early-to-mid stages of the trajectory.

\paragraph{Skill Selection and Visual Grounding.}


The core analysis reveals that the router's failures are not random but instead exhibit a clear pattern of visual grounding deficits. 
%
The \textit{Landmark Detection} expert is severely under-utilized, being the optimal skill in 76\% (13 of 17) of the analyzed failures, yet the router correctly selected it in only one instance. Instead of activating the \textit{Landmark} expert to identify specific objects—such as a curio cabinet or bed—the router frequently defaulted to coarser navigational skills like \textit{Region Identification} (6 cases) or \textit{Direction Adjustment} (5 cases). This mismatch indicates that while the router successfully interprets the linguistic instruction to move or change areas, it fails to bind the instruction to the specific visual target required to trigger the \textit{Landmark} expert. This confirms that the VLM experts still struggle to bind visual object references effectively even with Temporal Reordering in place, suggesting that the primary error source is visual grounding rather than high-level planning logic.

\paragraph{Failure Modes.}
Qualitative analysis identifies several specific mechanisms by which these failures manifest. 
A frequent cause of early failure is initial viewpoint misalignment, where the agent starts with an orientation that does not align with the instruction; this initial drift often propagates downstream as the router struggles to recover from an incorrect starting heading. 

Hallucinations also play a significant role in derailing routing. For example, the agent generated a subgoal to ``Position yourself beside the glass case with the dolls' when the instruction simply stated ``Enter the bedroom,' illustrating that the action router can create irrelevant descriptions grounded in neither the text nor the view which subsequently propagate incorrect skill choices. 

Furthermore, even when the router selects the correct skill, execution can still fail due to controller limitations. For instance, valid \textit{Direction} transitions failed when multiple direction subgoals were stacked (e.g., ``Turn left. Turn right. Turn left again.''), indicating that the skill-conditioned controllers themselves can become a bottleneck during complex sequences. Finally, the router occasionally fails to generate necessary ``Stop'' subgoals or struggles with ambiguous spatial prepositions, leading to incorrect termination points.

In summary, the analysis demonstrates that the routing mechanism is not acting as a random black box. It exhibits strong instruction understanding capabilities—often sequencing subgoals correctly in text—but acts as a bottleneck when it relies on textual priors over visual evidence. The frequent substitution of \textit{Region/Direction} skills for \textit{Landmark} skills highlights that improving the router's ability to attend to visual object cues is the most critical path for future performance gains.

\subsection{SkillNav is Not Data Augmentation}\label{sec:appendix_skill_mixed}


We clarify that SkillNav is not a data augmentation method, even though it uses synthetic skill-specific data for training individual skill agents. The role of synthetic data in our framework is to enable explicit specialization of atomic skills, not to improve a single monolithic policy. The core contribution of SkillNav is the explicit decomposition of navigation into atomic skills and the dynamic recomposition of these skills at inference time through a structured MoE routing mechanism. This is fundamentally different from data augmentation, which still relies on a single end-to-end policy without modular execution or dynamic expert selection.


To empirically address this concern, we conducted controlled experiments (Table \ref{tab:duet_eval_logs} on GSA-R2R and Table \ref{tab:performance_navnuance} on NavNuances) comparing SkillNav against a ``Mixed Skills'' baseline, a single monolithic DUET agent fine-tuned on the union of all synthetic skill data used in SkillNav.

If the performance gains were driven solely by data augmentation, the ``Mixed Skills'' agent should match SkillNav’s performance. However, our results show the opposite, confirming that structural decomposition (MoE), not data volume, is the driver of improvement.

As shown in Table \ref{tab:duet_eval_logs}, the ``Mixed Skills'' agent achieves an SR of $54.26\%$ on \textbf{Test-N-Scene}, which is statistically indistinguishable from the baseline ScaleVLN ($54.50\%$). In stark contrast, SkillNav achieves $56.66\%$ SR ($+2.4\%$ absolute gain). This gap confirms that simply exposing a model to compositional data is insufficient. The model requires the MoE architecture to dynamically route sub-problems to specialized experts rather than forcing a single network to learn conflicting behaviors.

Table \ref{tab:performance_navnuance} illustrates why the monolithic approach fails and why our MoE method aligns with the motivation of compositional reasoning.
\begin{itemize}
\item \textbf{Task Interference in Monolithic Models:} The ``Mixed Skills'' agent attempts to learn all behaviors simultaneously, resulting in diluted performance. For example, on Direction Change (DC), the Mixed model scores $66.84\%$.
\item \textbf{Superiority of Specialized Experts:} The specialized ``Direction Adjustment'' expert achieves $70.81\%$ on DC, significantly outperforming the mixed model. Similarly, the ``Vertical Movement'' expert achieves $87.65\%$ SR on VM tasks, compared to $84.11\%$ for the mixed model.
\end{itemize}
This demonstrates that SkillNav functions as a true MoE system: it leverages agents that have ``overfit'' to specific atomic skills (high performance in their domain, lower in others) and uses the router to compose them. A data augmentation approach cannot replicate this sharpness in skill execution.

The Role of Temporal Reordering in MoE Finally, we clarify that the Temporal Reordering Module is not an auxiliary processing step but a structural prerequisite for the MoE router. As shown in our ablations, without reordering, the router struggles to simultaneously decouple temporal segmentation from skill classification. The reordering module provides the necessary planning scaffold (an ordered list of subgoals), allowing the router to focus exclusively on visual grounding and expert selection.

\begin{table*}[h]
\centering
\caption{GSA-R2R DUET agent performance (SR/SPL) on Test-R-Basic, Test-N-Basic, and Test-N-Scene. Ident. = Identification. It shows that data augmentation alone fails.}
\label{tab:duet_eval_logs}
\begin{tabular}{c|cc|cc|cc}
\toprule
\multirow{2}{*}{\textbf{Agent}}
& \multicolumn{2}{c|}{\textbf{Test-R-Basic}}
& \multicolumn{2}{c|}{\textbf{Test-N-Basic}}
& \multicolumn{2}{c}{\textbf{Test-N-Scene}} \\
\cmidrule(lr){2-3} \cmidrule(lr){4-5} \cmidrule(lr){6-7}
& \textbf{SR} & \textbf{SPL}
& \textbf{SR} & \textbf{SPL}
& \textbf{SR} & \textbf{SPL} \\
\midrule

ScaleVLN & $78.12$ & $66.70$ & $69.13$ & $57.16$ & $54.50$ & $42.70$ \\
Mixed Skills & $77.60$ & $66.76$ & $70.23$ & $59.32$ & $54.26$ & $43.68$ \\

\midrule


Direction Adjustment  & $78.80$ & $68.40$ & $70.88$ & $59.53$ & $53.48$ & $42.22$ \\
Vertical Movement & $78.42$ & $67.52$ & $70.58$ & $59.22$ & $54.57$ & $43.78$ \\
Landmark Detection & $78.56$ & $67.84$ & $70.41$ & $59.65$ & $54.27$ & $44.77$ \\
Area and Region Ident. & $78.22$ & $68.24$ & $71.27$ & $60.81$ & $54.60$ & $44.93$ \\
Stop and Pause & $78.38$ & $67.93$ & $70.03$ & $58.77$ & $53.44$ & $42.02$ \\

\midrule

SkillNav  & $\mathbf{78.83}$ & $\mathbf{68.88}$ & $\mathbf{71.58}$ & $\mathbf{61.34}$ & $\mathbf{56.66}$ & $\mathbf{47.96}$ \\

\bottomrule
\end{tabular}
\end{table*}

\subsection{Novel Instruction Handling Mechanism}


We clarify that the improvement on Test-N-Scene does not come from the Temporal Reordering module alone, nor is it due to any scene-specific instruction handling built into the model. Instead, the gain arises from the joint effect of structured temporal planning and Mixture-of-Skills execution, which together enable SkillNav to better handle complex, long-horizon, and stylistically diverse instructions.

The Temporal Reordering Module plays a structural role inside the routing pipeline. It converts free-form instructions into an explicitly ordered subgoal sequence and supplies this sequence \textbf{only to the action router for subgoal localization}. Importantly, it does not execute actions, nor does it provide any extra semantic supervision to the skill agents. All skill-based agents still operate solely on the original full instruction and visual observation. Therefore, temporal reordering alone cannot account for the navigation improvement.

The action router is responsible for mapping each localized subgoal to the appropriate specialized skill agent. Without the MoE routing, the system cannot translate the reordered plan into skill-conditioned execution. This is directly supported by our ablation results: when routing is random or misaligned, performance drops even with temporal reordering enabled. Conversely, when both reordering and structured MoE routing are enabled, performance becomes stable and significantly improves, especially on Test-N-Scene.


Scene-style instructions in Test-N-Scene are typically written in a travel-speaker or tour-guide style, with rich descriptive language, delayed action cues, and loosely structured narration. For example, instructions often follow a pattern such as: “Alright folks, let’s take a gentle stroll down these beautiful marble steps… keep going straight ahead… you’ll notice the stone pillars on your left… right beyond them, the large double doors… that’s where we’ll pause.”

In summary, the performance gains on Test-N-Scene cannot be attributed to the reordering module alone. Temporal reordering reduces ambiguity in planning, while the MoE router is essential for translating each subgoal into the correct expert action. The improvement comes from their tight integration, validating that the gain is fundamentally due to structured compositional reasoning via skill recomposition, rather than instruction parsing alone.


\paragraph{Qualitative Example: Routing on a Scene-Style Instruction.}

To illustrate this synergy, consider a typical narrative instruction from Test-N-Scene:

“Alright folks, let’s take a gentle stroll down these beautiful marble steps here. Keep going straight ahead, and you’ll notice those magnificent stone pillars on your left. Just keep moving forward until you see the wooden confessionals and, right beyond them, those impressive large double doors. That’s where we’ll pause.”

The Temporal Reordering module first distills this narrative into discrete subgoals (e.g., “Begin walking down the marble steps,” “Observe the stone pillars,” etc.). However, SkillNav does not rigidly execute this list; instead, it dynamically routes control based on the visual environment.

\section{R2R Performance Ceiling and GSA-R2R Generalization}\label{sec:r2r_gsa_gap_appendix}

The smaller improvement on R2R and larger improvement on GSA-R2R come from two factors: limited headroom on R2R and a stronger distribution shift in GSA-R2R. R2R is approaching a performance ceiling. As reported in the original R2R benchmark, human navigators achieve an SR of $0.86$ and SPL of $0.76$ on the R2R test set~\citep{anderson2018vision}. SkillNav reaches an SR of $0.84$ and SPL of $0.77$ (Table~\ref{tab:r2r_human_upper_bound}), effectively matching human-level path efficiency. As a result, the remaining headroom for large absolute gains on this benchmark is limited.

\begin{table}[h]
\centering
\caption{Performance comparison with human navigation on the R2R test set.}
\label{tab:r2r_human_upper_bound}
\resizebox{0.86\linewidth}{!}{
\begin{tabular}{lcc}
\toprule
\textbf{Method / Navigator} & \textbf{SR $\uparrow$} & \textbf{SPL $\uparrow$} \\
\midrule
Human (Upper Bound)~\citep{anderson2018vision} & $0.86$ & $0.76$ \\
SkillNav (Ours) & $0.84$ & $0.77$ \\
\bottomrule
\end{tabular}
}
\end{table}

By contrast, SkillNav is designed for generalization rather than only in-domain performance. GSA-R2R contains colloquial, temporally complex, and dialogue-style instructions. In this setting, the specialized atomic skills, such as Landmark Detection and Area/Region Identification, act as robust anchors for execution. A monolithic agent can be distracted by conversational fillers in a travel-guide-style instruction, whereas the VLM-based router can identify the core navigation intent and route it to the appropriate specialized expert. This explains why the improvement scale is larger on GSA-R2R: the benchmark stresses linguistic and environmental generalization, while R2R primarily measures performance in a near-saturated in-domain setting.

\section{Broader Generalization on RxR-English}\label{sec:appendix_rxr_generalization}

We evaluate SkillNav zero-shot on RxR-English~\cite{ku_room-across-room_2020} to test generalization to longer and denser instructions than R2R. RxR emphasizes fine-grained path following and contains longer trajectories, so nDTW is especially informative because it penalizes deviations from the intended path rather than only checking final goal success. All models use the same image features and hyperparameter settings as their corresponding R2R/GSA-R2R configurations, with no RxR fine-tuning.

These results in Figure~\ref{fig:rxr_generalization} complement GSA-R2R. REVERIE primarily tests remote object grounding, R4R emphasizes extended path composition, and RxR emphasizes dense path adherence. The RxR gains therefore support the claim that the modular decomposition improves instruction-following fidelity beyond the main R2R/GSA-R2R setting.

\section{Efficiency Analysis}\label{sec:router_runtime_analysis_appendix}

All experiments in efficiency analysis in Section~\ref{sec:efficiency_analysis} run on NVIDIA A6000. For the inference cost in Table~\ref{tab:runtime}, the number of predictions is $14,400$ for Test-R-Basic and $9,000$ for Test-N-Basic.
For fairness, MapGPT is re-implemented with Qwen2.5-VL-7B-Instruct. 


To isolate router overhead, we disable Temporal Reordering and compare three VLM routers on GSA-R2R. As shown in Table~\ref{tab:router_runtime_analysis_appendix}, Qwen2.5-VL-7B-Instruct runs fastest at $18.12$ seconds per sample, while GLM-4.1V-9B-Thinking and InternVL3.5-8B increase per-sample latency to $55.99$ and $79.16$ seconds, respectively. Heavier routers therefore incur substantial runtime cost without material gains in routing robustness.

\begin{table}[h]
\centering
\caption{Runtime cost of VLM routers on GSA-R2R with Temporal Reordering disabled. Lower is better.}
\label{tab:router_runtime_analysis_appendix}
\resizebox{\linewidth}{!}{
\begin{tabular}{lcc}
\toprule
\textbf{Router} & \textbf{Inf. Time/Step (s)} & \textbf{Inf. Time/Case (s)} \\
\midrule
Qwen2.5-VL-7B-Instruct & $1.57$ & $18.12$ \\
GLM-4.1V-9B-Thinking & $4.74$ & $55.99$ \\
InternVL3.5-8B & $6.54$ & $79.16$ \\
\bottomrule
\end{tabular}
}
\end{table}

\begin{table}[t]
\centering
\caption{Real-time Inference Latency and Deployment Token Consumption Analysis.}
\label{tab:efficiency_routers_deployment_costs}
\resizebox{\linewidth}{!}{
\begin{tabular}{lccc}
\toprule
\textbf{Model (Router)} & \textbf{\makecell[l]{\# Calls\\ /Case}} & \textbf{\makecell[l]{Tokens\\ /Case}} & \textbf{\makecell[l]{Inf. Time (s)\\ /Case}} \\
\midrule
NavGPT (GPT-4o-mini) & 14.3  & 32,887 & 42.00 \\
FlexVLN (GPT-4o-mini) & 6.92  & 4,059  & 20.40 \\
\textbf{SkillNav (GPT-4o)} & 10.26 & 10,946 & \textbf{9.69} \\
\bottomrule
\end{tabular}
}
\end{table}

Furthermore, while SkillNav incorporates approximately $50\%$ more reasoning steps and richer historical visual context than FlexVLN to ensure robust long-term reasoning, it maintains superior operational efficiency and an optimized token-to-performance ratio as detailed in Table~\ref{tab:efficiency_routers_deployment_costs}.

\subsection{Why the skill-based agents are not VLM-based models}

The SkillNav framework itself is model-agnostic and only assumes the availability of specialized skill executors. In principle, these skill executors could be instantiated by any policy model. In our implementation, however, we deliberately use pretrained supervised VLN backbones rather than VLMs for skill execution, based on both effectiveness and efficiency considerations.


From an effectiveness perspective, existing LLM or VLM-based VLN agents such as MapGPT, NavGPT-2, VLN-R1, and DiscussNav consistently lag behind supervised VLN models on R2R, and more importantly, they exhibit unstable generalization on GSA-R2R. This shows that directly using MLLMs as low-level action policies is still unreliable for precise and robust navigation, especially in unseen environments.




From an efficiency perspective, our analysis in Section~\ref{sec:efficiency_analysis} already shows that using a single MLLM as the router makes SkillNav about $50\times$ slower than a fully supervised VLN model. The throughput ablation in Table~\ref{tab:router_runtime_analysis_appendix} further quantifies this cost across different routers: even among zero-shot VLMs, per-sample inference ranges from 18.12 to 79.16 seconds, indicating that the router alone is already the dominant computational bottleneck.
If we further replace each low-level skill executor with an MLLM or VLM, the system would require multiple MLLM inferences per navigation step, which would introduce another order-of-magnitude slowdown. This would make the system impractical for real-time or large-scale deployment in real-world embodied settings.


In summary, SkillNav is a general, model-agnostic, skill-based framework that is fully compatible with the MLLM paradigm at the routing and planning level. However, for executing atomic navigation skills, we intentionally prefer pretrained, compact VLN backbones over VLMs, because they provide significantly better accuracy, efficiency, and spatial control reliability for low-level navigation.

\section{Evidence of Supervised Agents' Memorization}\label{sec:appendix_supervised_agents_memorization}

While recent methods like ScaleVLN and SRDF have indeed narrowed the performance gap on standard splits (val-seen vs. val-unseen), this improvement does not reflect true generalization. Instead, our analysis suggests these methods rely on linguistic memorization (due to high distribution overlap) and task-specific overfitting (via augmentation construction), causing them to fail in rigorous generalization benchmarks like GSA-R2R.

\paragraph{Linguistic Memorization due to Distributional Homogeneity.}

The ``unseen'' environments in R2R benchmarks are not linguistically novel. Our analysis of the R2R instruction vocabulary reveals that the distribution of instructions in validation splits is statistically indistinguishable from the training distribution.
There is a Jaccard \textbf{similarity score of $0.9992$} between instruction sets, with effectively perfect token coverage.
The Cosine similarity of token frequency vectors is $1.0$, and the Jensen–Shannon divergence is negligible ($8.2 \times 10^{-6}$). This implies that the ``narrowing gap'' is likely due to models memorizing these static linguistic patterns, which remain constant across splits, rather than learning to ground instructions in truly unseen visual environments.

\paragraph{Failure in Robust Generalization Benchmarks (GSA-R2R).}

To test if the performance gap reduction translates to real-world robustness, we evaluated these methods on GSA-R2R, a benchmark designed to test generalization beyond standard splits.

Despite high scores on standard R2R, both ScaleVLN and SRDF exhibit significant performance drops in GSA-R2R testing.

Crucially, this failure persists even when the augmentation data includes scenes that were technically ``seen'' during training. This indicates that their success on R2R is brittle and relies on specific dataset biases that do not hold in broader generalization scenarios.

\paragraph{Methodological Memorization via Data Construction.}

Finally, we argue that the definition of ``memorization'' extends to the training methodology.

Methods like SRDF require constructing specific augmentation data tailored for each new task.

This reliance on constructing task-specific priors acts as a form of structural memorization. The model is not generalizing to a new task zero-shot; it is overfitting to the constraints provided by the engineered augmentation data.

\section{LLM Usage}
We used LLM-based tools for polishing grammar and aiding in writing. In addition, we utilize a LLM to generate synthetic instructions for skill-specific datasets as described in Section~\ref{sec:data_synthesis_agent_train}.
Moreover, LLMs and VLMs serve as our temporal reordering module and action router in Section~\ref{sec:temporal_reordering}.

\section{LLM and VLM Prompts}\label{sec:prompts}

In this section, we provide the prompts used in data construction and all components of SkillNav.

\subsection{Prompts for Skill-specific Data Synthesis}\label{sec:prompt_data}

To generate skill-focused instruction, we feed the observation sequence of each candidate trajectory into GPT-4o
with the structured prompt, in Listing~\ref{lst:temporal_data_Synthesis} and Listing~\ref{lst:atomic_data_Synthesis}. Both of prompts are tailored for GPT-4o.



\paragraph{Temporal Order Planning Skill Data Construction.}
The detailed prompt for Temporal Order Planning Skill data construction can be seen in Listing~\ref{lst:temporal_data_Synthesis}.

\begin{figure*}[!ht]
\centering
\captionsetup{type=listing}
\begin{lstlisting}[caption={Prompt used for Temporal Order Planning Skill-specific Data Synthesis}, label={lst:temporal_data_Synthesis}]
You are an expert in Vision-and-Language Navigation (VLN) and Language.

<Task>
Your task is to write natural, human-like navigation instructions based on a sequence of visual observations from an indoor environment.

<Instruction Guidelines>
- Do not use explicit temporal markers like ``then'', ``next'', ``before'', or ``after''.
- Imply sequence using spatial or contextual phrasing instead.
- Include only essential visual cues, such as layout, furniture, doorways, or notable landmarks that help clarify the path.
- Avoid over-descriptive or decorative language (e.g., ``intricate stonework'', ``high ceiling'').
- Keep the instruction fluent, intuitive, and helpful, like someone casually guiding a friend through a space.
- Keep it concise and comparable in length to a temporal-based instruction.

<Visual Reasoning Process>
Analyze each frame in the visual sequence. Focus on:
- Movement across spaces
- Transitions (e.g., turns, room entries)
- Orientation shifts
- Key visible cues needed to navigate the path

<Instruction Output>
Once you've analyzed the path:
- Write a fluent, natural-sounding instruction describing the full trajectory.
- Do **not** include reasoning steps.
- Output **only** the final instruction.

<Example Chain-of-Thought>
- 1st Frame:
    - The agent is inside a narrow wooden hallway-like space.
    - The doorway directly ahead leads to a brighter area.

- 2nd Frame:
    - The agent is almost at the threshold of the doorway.
    - You can see the hallway plant and the open area outside.

- 3rd Frame:
    - The agent is now fully outside the room, looking into a wide open space.
    - There's a visible bedroom to the left, and the plant in the yellow pot is to the right, indicating the agent has made a hard left turn.

- 4th Frame:
    - The agent is now facing a doorway to a bedroom on the left side.
    - The bed is partially visible inside.

- 5th Frame:
    - The agent has entered the room and is facing a window.
    - The position suggests the agent took one step inside and then stopped.

---

<Trajectory Images>
``{path_images}''


\end{lstlisting}
\end{figure*}

\paragraph{Atomic Skills Data Construction.}
The 5 atomic skills in VLN share the same prompt (in Listing~\ref{lst:atomic_data_Synthesis}) for their skill-specific data synthesis.

\begin{figure*}[!ht]
\centering
\captionsetup{type=listing}
\begin{lstlisting}[caption={Prompt used for Atomic Skill-specific Data Synthesis}, label={lst:atomic_data_Synthesis}]
You are an expert in Vision-and-Language Navigation (VLN) and Language.

<Task>
- Generate a **single** natural-language instruction that guides an agent through the scene.

<Input>
- A visual sequence (an ordered list of images)
- A specific navigation skill to emphasize

<Requirements>
- The instruction should describe what the agent does across the image sequence (e.g., move, climb, pause).
- Ground the instruction in **visible cues**, such as layout, objects, stairs, doorways, lighting, or orientation.
- Emphasize the given **target skill** (e.g., "Direction Adjustment", "Vertical Movement", etc.), while naturally incorporating other relevant details as needed.
- The output must be a **single sentence**, written in fluent, natural language (no lists, quotes, or symbols).
- Instruction length should be **20-30 words** (aim for ~25).
- Do **not** include explanations, reasoning steps, or metadata output only the instruction itself.

<Available Skills>
{Direction Adjustment, Vertical Movement, Stop and Pause, Landmark Detection, Area and Region Identification}

<Skill Definitions>
- **Direction Adjustment**: Involves turning or changing heading. Look for instructions like ``turn left'', ``go back'', or ``face the hallway''. Used when the agent needs to rotate or reorient without necessarily changing position.

- **Vertical Movement**: Involves moving across floors or elevation changes. Triggered by terms like ``go upstairs'', ``down the stairs'', or ``take the elevator''. Watch for floor changes in visuals or references to vertical navigation.

- **Stop and Pause**: Involves coming to a full stop at a defined point. Use lighter-weight verbs such as pause, wait, and stand, when the stop happens in the middle of sequence (e.g., ``pause by the red sofa''). Use stronger, more terminal verbs like stop and come to a stop for the final action or true endpoint (e.g., ``stop at the glass doors''). This distinction helps the agent decide whether to hold briefly or end its navigation.

- **Landmark Detection**: Requires identifying and responding to specific objects or features in the environment. Triggered by mentions of visible items like ``lamp'', ``chair'', ``red sofa'', ``painting''. Used when object recognition is necessary to proceed or confirm position.

- **Area and Region Identification**: Involves recognizing or transitioning between distinct spaces or rooms. Triggered by mentions like ``enter the kitchen'', ``in the bedroom'', ``exit hallway''. Requires understanding of semantic regions based on context or appearance.


<Output Format>
Return only the instruction sentence. Do not include tags, labels, or formatting.

---

<Trajectory Images>
``{path_images}''

<Focused Skill>
``{skill_name}''

\end{lstlisting}
\end{figure*}

\subsection{Prompt for Temporal Reordering Module}

The Temporal Order Module only takes the original natural language instruction
as input. It applies the instruction reordering prompt to turn navigation instructions into subgoals $I_{\text{reorder}}$. 
The prompt is shown in Listing~\ref{lst:temporal_prompt}, utilizing GPT-4o as the generation model.

\begin{figure*}[!ht]
\centering
\captionsetup{type=listing}
\begin{lstlisting}[caption={Prompt used for Temporal Reordering}, label={lst:temporal_prompt}]
You are an expert at converting natural language navigation instructions into detailed, logically ordered sub-instructions for agents.

<Task>
- Break down instructions into a sequence of minimal, goal-directed steps.
- Make all implicit temporal or spatial relationships explicit.
- Preserve execution order by reconstructing intermediate actions that are implied, not directly stated.

<Logic Rules>
- (A) --> [after / then / once / as soon as] --> (B): Do A fully, then B.
- (B) --> [before] --> (A): Move toward A, then perform B at a point prior.
- (A) --> [until] --> (B): Continue A until B is reached.
- Avoid ``then'', ``before'', ``until'', ``once'' etc. in the output.

<Formatting Rules>
- Single sentence, steps separated by periods.
- Each step must be minimal, concrete, and goal-focused.

<Examples>
**Example 1:**
Instruction: ``Turn around and walk down the stairs. Stop once you get down them.''
Output:
Turn around. Walk down the stairs. Stop at the bottom of the stairs.

**Example 2:**
Instruction: ``Walk toward the dining room but turn left before entering it and go into the open area.''
Output:
Walk toward the dining room. Stop at the entrance. Turn left. Enter the open area.

**Example 3:**
Instruction: ``After you leave the laundry room, make a left in the hallway, and go to the bedroom straight ahead. When you are in the doorway of the room go to the doorway of the closet on the left and wait.''
Output:
Exit the laundry room. Turn left in the hallway. Walk to the bedroom straight ahead. Enter the doorway of the bedroom. Go to the doorway of the closet on the left. Wait there.

**Example 4:**
Instruction: ``Start moving forward down the corridor. You will pass offices on your left and right. Keep going down the hallway until you get to an exit sign on your right and what looks like some lockers in front of you. There will also be a brown door with an exit sign above it in front of you.''
Output:
Start moving forward down the corridor. Pass the offices on your left and right. Continue walking down the hallway. Reach the exit sign on your right and the lockers in front of you. Stop in front of the brown door with the exit sign above it.

---

<Original Instruction>:
``{instruction}''


\end{lstlisting}
\end{figure*}

\subsection{Prompts for Action Router}\label{sec:appendix_prompt_action_router}
The Action Router dynamically selects the most suitable agent at each time step, which can be structured into two distinct reasoning phases:
Phase 1 Subgoal Localizer and Phase 2 Skill Router. We provide the detailed prompt for the two phases, respectively. They can be used for either Qwen2.5-VL-7B-Instruct
or GLM-4.1V-Thinking-9B.
.

\paragraph{Subgoal Localizer.}
The Subgoal Localizer identifies the next subgoal to be executed for the current time step and outputs the corresponding reasoning trace. Listing~\ref{lst:subgoal_localizer} contains the prompt for the subgoal localizer, which takes all reorder subgoals, the previously executed subgoals, and the prior selected viewpoints as input.

\begin{figure*}[!ht]
\centering
\captionsetup{type=listing}
\begin{lstlisting}[caption={Prompt used for Subgoal Localizer in Action Router}, label=
{lst:subgoal_localizer}]
You are a visual reasoning assistant for indoor navigation.
<Task>:
Your task is to analyze a list of previously observed images and a natural language instruction.
Determine which parts of the instruction have already been completed, and return the next step to be executed.
<Response Rules>
Your response must:
- Return the next action using *exact phrasing* from the reordered instruction (no paraphrasing).
- Match the sub-instruction to the visual context from previous images.
- If the goal (e.g., pool table) has clearly been reached, return the final sub-instruction.
- If *all* sub-instructions have been completed based on the visual path, do not return anything further. Stop reasoning.
- If the final destination has been reached and the last step is a positional or waiting action (e.g., ``wait there'', ``step to the left''), return that as the next step.
- You must reason about whether the agent is already at the destination.
- If the current image shows the goal destination (e.g., inside the room with the pool table, or inside the open doorway), and the instruction contains a final step like ``wait'' or ``adjust your position'', that is the next sub-instruction.
---
Use the following reasoning strategy to determine what to do next:
<Step-by-Step Reasoning Instructions>:
1. Decompose the instruction into sub-instructions.
- Break the full instruction into smaller steps. Each sentence or clause typically represents one step.
- Example:
    - Original: ``At the bottom of the stairs, go through the nearest archway to your left. Head straight until you enter the room with a pool table. Step slightly to the left to get out of the way.''
    - Decomposed:
        - ``At the bottom of the stairs, go through the nearest archway to your left.''
        - ``Head straight until you enter the room with a pool table.''
        - ``Step slightly to the left to get out of the way.''
2. Use the previous sub-instruction list to identify completed steps.
- Do not reissue any previously executed sub-instructions.
- Compare upcoming steps against what may have been visually completed, even if not explicitly executed one-by-one.
3. Analyze the sequence of previous viewpoint images.
- Use visual context to infer if *multiple* sub-instructions have been completed in a single transition.
- If image progression clearly shows the agent has already bypassed an intermediate area or reached a later goal, mark those steps as implicitly complete.
4. Evaluate remaining sub-instructions for completion.
- If the current image shows the agent at or beyond the target of a sub-instruction, that step can be considered completed.
- If the current image shows the agent inside the goal location and only a final positional instruction remains (e.g., ``Step slightly to the left''), return that final instruction.
5. Select the next uncompleted sub-instruction that is visually and contextually justified.
- Use exact wording from the original instruction.
- Do not return instructions that the agent already visually fulfilled, even if they were skipped.
6. Output the result in the following JSON format:
{
"Sub-instruction to be executed": "<exact next instruction clause>",
"Reasoning": "<why this is the next step based on image sequence>"
}
CHECKPOINT:
If multiple sub-instructions were completed based on a single or continuous image segment, skip them and jump to the next logical, visually unfulfilled step.
---

Now, using the instruction and the visual history, identify the next step.
IMPORTANT: Your response must be a valid JSON object without any surrounding text, code blocks, or explanations.
Do not include markdown formatting like ```json or ```.

<Original Whole Instruction>:
``{instruction}''
<Previous Sub-Instructions>:
``{previous_sub_instructions}''
<Previous Viewpoint Images>:

\end{lstlisting}    
\end{figure*}

\paragraph{Skill Router.}
The skill router determines which skill-based agent is most appropriate for executing the selected subgoal among the 5 skill-based agents. Besides, it receives the original instruction as contextual input to capture additional linguistic cues such as verbs and spatial references. It also uses the
reasoning trace from the subgoal localizer to enhance its understanding of the current subgoal. The whole process is displayed in Listing ~\ref{lst:skill_router}.

\begin{figure*}[!ht]
    \centering
    \begin{lstlisting}[caption={Prompt used for Skill Router in Action Router}, label={lst:skill_router}]
You are a visual reasoning assistant for indoor navigation.

<Available Skills>:
[``Direction Adjustment'', ``Vertical Movement'', ``Stop and Pause'', ``Landmark Detection'', ``Area and Region Identification'']

<Skills Explanation>:
- Direction Adjustment:
Involves turning or changing heading. Look for instructions like ``turn left'', ``go back'', or ``face the hallway''. Used when the agent needs to rotate or reorient without necessarily changing position.
- Vertical Movement:
Involves moving across floors or elevation changes. Triggered by terms like ``go upstairs'', ``down the stairs'', or ``take the elevator''. Watch for floor changes in visuals or references to vertical navigation.
- Stop and Pause:
Involves stopping at a specific location. Triggered by instructions like ``stop'', ``wait'', or ``stand in front of''. Used when the endpoint or a mid-action pause is important.
- Landmark Detection:
Requires identifying and responding to specific objects or features in the environment. Triggered by mentions of visible items like ``lamp'', ``chair'', ``red sofa'', ``painting''. Used when object recognition is necessary to proceed or confirm position.
- Area and Region Identification:
Involves recognizing or transitioning between distinct spaces or rooms. Triggered by mentions like ``enter the kitchen'', ``in the bedroom'', ``exit hallway''. Requires understanding of semantic regions based on context or appearance.

<Task>:
1. Read and understand the sub-instruction to be executed.
2. Use the reasoning explanation to infer what skills are likely required to carry out that sub-instruction.
3. Choose the top 1 skill that is most relevant to the sub-instruction.

<Input>:
You will be given:
- The original full navigation instruction.
- The sub-instruction that should be executed next, based on reasoning.
- A reasoning explanation derived from the visual history and instruction.

Output exactly **one skill name** from the above list.
Do not provide explanations or additional text.

<Output Format>:
*****SKILL_NAME*****

<Example>
Original Whole Instruction: ``At the bottom of the stairs, go through the nearest archway to your left. Head straight until you enter the room with a pool table. Step slightly to the left to get out of the way.''

Sub-instruction to be executed for next step: ``Head straight until you enter the room with a pool table.''

Reasoning based on previous viewpoints path and original instruction: The agent appears to be just outside the archway. The next step is likely to involve entering the archway and preparing to head straight.

Expected Output:
*****Landmark Detection*****

---

<Reordered Whole Instruction>: 
``{full_instruction}''

Sub-instruction to be executed for next step: 
``{sub_instruction}''

<Reasoning based on previous viewpoints path and original instruction>: 
``{reasoning}''
        
    \end{lstlisting}
\end{figure*}

Execution begins when the router detects the stair structure and assigns the current subgoal to the Vertical Movement skill. Crucially, this agent receives the original full instruction. Upon activation, it parses the complete narrative to locate and attend to the specific phrase “down these beautiful marble steps,” grounding this text to the visual descent while ignoring later landmark descriptions.

Once the agent descends and aligns with the corridor, the router shifts control to the Direction Adjustment skill. This agent processes the full instruction to isolate the directive “keep going straight ahead,” executing the forward movement. Subsequently, when the router activates the Landmark Detection skill for the ``stone pillars' subgoal, the agent again references the original text. It attends to the detailed description—“magnificent stone pillars on your left”—to perform precise visual grounding, ensuring the agent identifies the specific object mentioned in the tour.

Finally, the router tracks the destination using the Stop and Pause skill. This agent verifies the visual presence of the doors against the instruction's final clause (“that’s where we’ll pause”) before terminating the episode. Throughout the trajectory, the router dictates which expert is active, but the experts themselves derive their visual-linguistic understanding directly from the rich, original instruction.

In summary, the improvements on Test-N-Scene are not attributable to better instruction parsing alone. While reordering provides necessary structure to long narratives, the MoE routing mechanism provides the essential visual grounding required to execute that structure. The advantage of SkillNav lies in this tight integration of planning and compositional execution.

\end{document}